\pdfoutput=1

\documentclass[11pt]{article}

\usepackage[final]{acl}

\usepackage{times}
\usepackage{latexsym}

\usepackage{amssymb}

\usepackage[T1]{fontenc}

\usepackage[utf8]{inputenc}

\usepackage{microtype}

\usepackage{inconsolata}

\usepackage{graphicx}
\usepackage{amsmath}
\usepackage{float} 
\usepackage{makecell}  
\usepackage{multirow}
\title{Do All Autoregressive Transformers Remember Facts the Same Way?\\A Cross-Architecture Analysis of Recall Mechanisms}

\author{
  \textbf{Minyeong Choe\textsuperscript{1}},
  \textbf{Haehyun Cho\textsuperscript{2}},
  \textbf{Changho Seo\textsuperscript{3}},
  \textbf{Hyunil Kim\textsuperscript{1}\thanks{Corresponding author.}}
\\
  \textsuperscript{1}Chosun University, 
  \textsuperscript{2}Soongsil University, 
  \textsuperscript{3}Kongju National University
\\
  \texttt{
  \textsuperscript{1}\{minyeong, hyunil\}@chosun.ac.kr,
  \textsuperscript{2}haehyun@ssu.ac.kr,
  \textsuperscript{3}chseo@kongju.ac.kr
  }
}

\begin{document}
\maketitle
\begin{abstract}
Understanding how Transformer-based language models store and retrieve factual associations is critical for improving interpretability and enabling targeted model editing.
Prior work, primarily on GPT-style models, has identified MLP modules in early layers as key contributors to factual recall. 
However, it remains unclear whether these findings generalize across different autoregressive architectures.
To address this, we conduct a comprehensive evaluation of factual recall across several models---including GPT, LLaMA, Qwen, and DeepSeek---analyzing where and how factual information is encoded and accessed.
Consequently, we find that Qwen-based models behave differently from previous patterns: attention modules in the earliest layers contribute more to factual recall than MLP modules. 
Our findings suggest that even within the autoregressive Transformer family, architectural variations can lead to fundamentally different mechanisms of factual recall.\footnote{The code and data are available at \url{https://github.com/ICT-Convergence-Security-Lab-Chosun/CAARM}}
\end{abstract}

\begin{figure*}[t]
  \includegraphics[width=\textwidth]{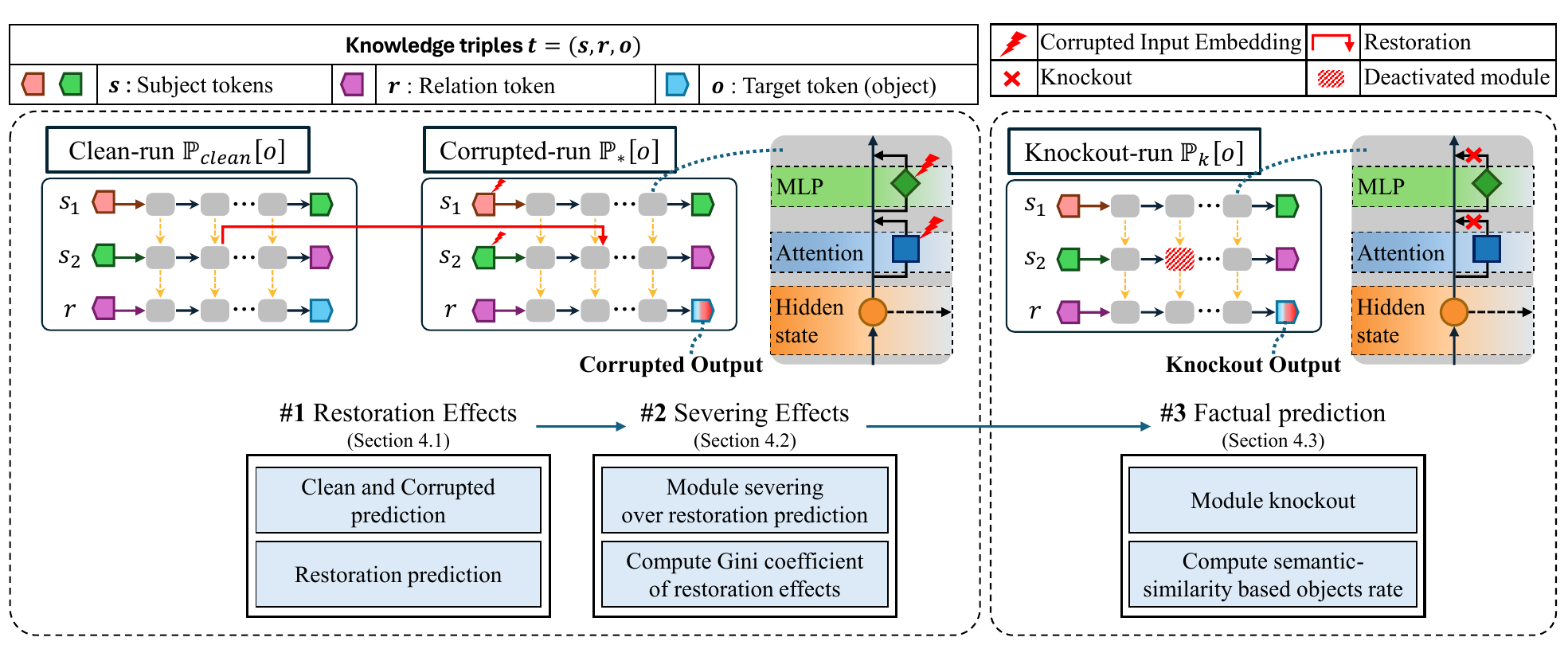}
  \caption{\textbf{Overview of Our Evaluation.} Given a sentence represented as knowledge triples of the form $t = (s, r, o)$, the pipeline proceeds in three stages: {\bf{\#1}} restoration effects, {\bf{\#2}} severing effects, and {\bf{\#3}} factual prediction. {\bf{\#1}} and {\bf{\#2}} are conducted using the indirect effect, which is computed from clean-run and corrupted-run. In contrast, {\bf{\#3}} is conducted on knockout-run. Each stage applies a distinct intervention strategy to evaluate the contribution of the Attention and MLP modules to factual association recall.}
  \label{Figure1}
\end{figure*}


\section{Introduction}
\label{s:intro}
Transformer-based language models 
are trained in an autoregressive manner, generating the next token sequentially based on previous tokens. These models exhibit strong language understanding due to their scale and self-attention mechanism~\cite{vaswani2017attention}. Beyond simple sentence generation, they internalize factual associations, typically expressed in knowledge triples of the form $t = (s, r, o)$, where $s$ is the subject, $r$ is the relation, and $o$ is the object, and effectively recall them during inference~\cite{petroni2019language}.

Factual association recall refers to a model’s ability to generate the correct object given a subject–relation prompt by leveraging internally stored factual knowledge. One line of research on factual recall investigates whether specific properties are embedded within the internal representations of language models, by tracing information flow through causal analysis~\cite{meng2022locating,meng2022mass,geva2023dissecting}. These approaches are widely used in applications such as knowledge editing to identify where factual associations are primarily stored and to enable targeted modifications, rather than editing the entire network.

However, these prior findings have focused only on GPT-family architectures. 
This observation raises the question of whether \textit{such findings generalize across different autoregressive architectures}. 
To address this gap, we conduct a comprehensive, layer-wise and module-wise (Attention vs. MLP) quantitative evaluation of factual association recall---examining where and how factual information is stored and retrieved across a range of autoregressive Transformer models, including GPT~\cite{radford2019language,brown2020language,gpt-j}, LLaMA~\cite{grattafiori2024llama}, Qwen~\cite{yang2024qwen2}, and DeepSeek~\cite{guo2025deepseek}~\footnote{Evaluation results are based on 17 models across target architectures; For brevity, we present results for four representative models in the main text, with the remaining results included in Appendix~\ref{Appendix_A}.}.
Our evaluation and analysis reveals \textit{notable architecture-specific differences}---particularly in the Qwen-based architectures---in where and how factual associations are recalled.

In our evaluation, we extend the experimental frameworks of \citet{meng2022locating} and \citet{geva2023dissecting}, applying causal tracing across a diverse set of autoregressive Transformer models to perform the following experiments, as showed in Figure~\ref{Figure1}.

\noindent
\textit{1. Restoration effects.}
We measure the change in inference probability between clean and corrupted inputs at the last subject token position, following causal tracing protocols from prior work. 
This allows us to analyze the sensitivity of model predictions across all layers and modules (Attention vs. MLP), providing insight into where factual associations are most affected by input corruption.

\noindent
\textit{2. Severing effects.}
We perform selective severing either the Attention or MLP module at each layer during inference. We then evaluate how much each module contributes to the model’s ability to generate the correct object in factual association prompts. 
In addition, we compute the Gini coefficient over inference contributions to assess how structurally localized the causal effects are within each module.


\noindent
\textit{3. Factual prediction.}
We refine the object prediction evaluation method used in \citet{geva2023dissecting}, which relies on string matching between predicted tokens and BM25-selected candidates. Noting its insensitivity to semantically equivalent outputs, we introduce a semantic similarity–based evaluation metric, which considers a predicted token correct if its embedding similarity to a candidate token exceeds a fixed threshold. 
This allows for more robust evaluation of model outputs, especially in architectures that generate more varied but semantically equivalent expressions.

\noindent
\textbf{Results.} 
This study highlights the following:

\noindent
1. Our evaluation confirms that, in GPT-based models, MLP modules in the early layers play a key role in storing factual associations. 
This finding supports prior work~\cite{meng2022locating, geva2023dissecting}, demonstrating that their conclusions are applicable to this class of architectures.

\noindent
2. However, we identify key factors indicating that Qwen-based models show larger changes in inference probability within the \textit{early Attention layers}, unlike GPT-based models. 
In Qwen-based models, factual associations are more concentrated in the Attention module than in the MLP. 
This finding is further supported by Gini coefficient analysis, which shows that the Attention modules are where most of the important effects are focused—highlighting their key role in factual recall in this model.


\noindent
3. Our semantic similarity–based evaluation further confirms the difference between GPT-based and Qwen-based models.
In Qwen-based models, Attention modules consistently contribute more to factual inference than MLPs, even when the model's outputs vary in wording.

These findings have practical implications for deploying Transformer models in real-world scenarios, \textit{where knowing which layers and modules store factual knowledge is important} for effective targeted model editing, reliable interpretability, and knowledge-intensive tasks.

\section{Notation of Autoregressive Transformer}
We begin by outlining the fundamental architecture of autoregressive Transformer models~\cite{vaswani2017attention}, before analyzing the recall mechanisms of factual associations in representative models such as GPT, LLaMA, Qwen, and DeepSeek. These models are designed such that each token can perform the Attention mechanism only over tokens that precede it.\\
\indent An autoregressive Transformer model $M : X \rightarrow Y$ operates over a vocabulary $V$, and given an input token sequence $x=[x_1, ..., x_T] \in X$, produces a probability distribution $y \in Y \subset \mathbb{R}^{|V|}$ for the next token. This distribution reflects the model's estimation of the next token given the preceding context.\\
\indent Internally, each token $x_i$ is mapped to a hidden state vector $h_i^{(l)}$ at layer $l$. The initial hidden state $h_i^{(emb)}$ is computed as the sum of the token embedding and positional embedding, as defined in Equation~\ref{equation1}.
\begin{equation}
  \label{equation1} 
  h_i^{(emb)}=emb(x_i)+pos(i)\in
  \mathbb{R}^{d_{model}}
\end{equation}
\indent The hidden state is iteratively updated through layers using Attention and MLP modules. At layer $l$, the hidden state of the $i$-th token is :
\begin{equation}
  \label{equation2}
  h_i^{(l)}=h_i^{(l-1)}+a_i^{(l)}+m_i^{(l)}
\end{equation}
\indent The Attention output $a_i^{(l)}$ incorporates representations from previous tokens via self-attention :
\begin{equation}
  \label{equation3}
  a_i^{(l)}=attn^{(l)}(h_1^{(l-1)}, ..., h_i^{(l-1)})
\end{equation}
\indent The MLP output applies a non-linear transformation to the residual input : 
\begin{equation}
  \label{equation4}
  m_i^{(l)}=W_{proj}^{(l)}\sigma(W_{fc}^{(l)}\gamma(a_i^{(l)}+h_i^{(l-1)}))
\end{equation}
\noindent where $\gamma$ denotes a normalization, and $\sigma$ is a non-linear activation.\\

\section{Structural Insights into Factual Recall} 
\label{section3}
\indent We conduct a quantitative analysis of how factual associations are stored across various autoregressive models through a unified evaluation. The results show that Qwen and DeepSeek exhibit patterns that differ from previously reported trends. Specifically, the early Attention layers corresponding to the position of the last subject token make significant contributions to the storage of factual associations. These findings suggest that architectural differences may influence the mechanisms underlying factual association storage.\\

\indent We frame our empirical study around the following research questions, and describe our approach.

\noindent {\bf{\#1.}} In various autoregressive Transformer models, which modules, layers, and token-position-specific activations contribute to the recall of specific factual associations, and to what extent do they influence the model’s output?
\begin{itemize}
  \item Following the causal tracing methodology introduced in ROME~\cite{meng2022locating}, we conduct {\bf{restoration effects}} and {\bf{severing effects}} experiments across a range of autoregressive Transformer models. Through these experiments, we quantitatively measure the Average Indirect Effect (AIE) of internal activations.
\end{itemize}
{\bf{\#2.}} Experiments conducted across various models reveal an inconsistency between the results of the restoration effects experiment and the severing effects experiment in the Qwen and DeepSeek models, raising the question: what factors might have caused this discrepancy?
\begin{itemize}
  \item To investigate the underlying cause of this discrepancy, we compute the \textbf{Gini coefficient} over the distribution of AIE values to quantify the degree of concentration across layers, thereby identifying which layers contribute most to factual association recall.
  \item Subsequently, we apply the severing effects experiment to the layers within the Attention modules of the Qwen and DeepSeek models that exhibit the highest AIE concentration. However, severing these layers does not result in a significant reduction in AIE. The inconsistency between the two experimental results is consistently observed.
\end{itemize}
{\bf{\#3.}} These findings suggest that the severing effects may have failed to sufficiently suppress the activations of the Attention module. Does this not indicate the need for a more precise intervention method?
\begin{itemize}
  \item Experimental results reveal that the severing effects method (as proposed in \citet{meng2022locating}) has a critical limitation in accurately reflecting the contribution to factual association recall, as it fails to sufficiently suppress the influence of the attention module.
  \item To address this limitation, we adopt the methodology proposed by~\citet{geva2023dissecting} and apply a \textbf{knockout} technique that directly blocks the output of each module. In addition, we evaluate factual prediction performance using the objects rate instead of AIE, incorporating semantic similarity by computing the cosine similarity between Sentence-BERT~\cite{SBERT} embeddings, rather than relying on exact string matching. Through this approach, we empirically confirm that the early layers of the Attention module in the Qwen and DeepSeek models contribute to the factual recall.
\end{itemize}

\indent We present a novel finding that, in Qwen-based models, the early layers of the Attention module contribute more to factual association recall at the last subject token position than the MLP module. This contrasts with prior work~\cite{meng2022locating,geva2023dissecting} that identifies the MLP as the primary site for factual association storage. Our result suggests that the localization of factual associations can shift depending on model architecture, highlighting the need for broader comparative studies across Transformer families.

\section{Evaluation Methods}
We build on and modify the implementations provided by~\citet{meng2022locating} and~\citet{geva2023dissecting} to suit our evaluation framework.

\noindent
{\bf{Model Selection.}} We evaluate a variety of autoregressive Transformer models, as summarized in Table~\ref{Table1}. 
Additional experimental targets and evaluation results are provided in the Appendix~\ref{Appendix_A}.

\noindent
{\bf{Dataset \& Prompting.}} We employ the \textsc{CounterFact} dataset~\cite{meng2022locating}, which is specifically designed to test factual associations stored in language models. 
This dataset consists of a diverse collection of prompts that encode factual associations.

To examine how each model processes the object in a factual association, we input only the subject and relation from a knowledge tuple $t=(subject, relation, object)$, excluding the object. This setup allows us to evaluate the model's ability to accurately predict the object based on the given context. For performance evaluation, we sample 100 factual sentences in which the object was successfully predicted, and apply the evaluation methodology described in the next subsection. Throughout all experiments, model layers are indexed from 0 to \(L-1\), and our analysis focuses on the last subject token position, which is critical for triggering factual recall within the model.


\begin{table}[t]
\centering
\begin{tabular}{ccc}
\Xhline{1.2pt}
\textbf{Model} & \textbf{\#Layers} & \textbf{\#Parameters} \\
\Xhline{0.8pt}
GPT-2-XL & 48 & 1.5B \\
\Xhline{0.2pt}
LLaMA-3.2-1B & 16 & 1B \\
\Xhline{0.2pt}
Qwen-2.5-1.5B & 28 & 1.5B \\
\Xhline{0.2pt}
\begin{tabular}[c]{@{}c@{}}DeepSeek-R1\\Distill-Qwen-1.5B\end{tabular} & 28 & 1.5B \\
\Xhline{1.2pt}
\end{tabular}
\caption{Architectural details of autoregressive Transformer models (B = Billion parameters).}
\label{Table1}
\end{table}

\subsection{Restoration Effects}

Our first experiment {\bf{\#1}} aims to quantitatively analyze how specific modules, layers, and token-position-specific activations in Transformer-based language models contribute to the storage and recall of factual associations.

The analysis assumes two baseline execution settings. The {\bf{clean-run}} executes the model on an uncorrupted input, yielding accurate factual predictions. In contrast, the {\bf{corrupted-run}} injects noise into the subject representation to degrade recall performance. Specifically, we perturb the subject embedding by adding noise with a magnitude of $v = 3\sigma_{sub}$, where $\sigma_{sub}$ is the standard deviation of subject embeddings collected from the dataset.

Based on this setup, we quantify the restoration effects by restoring the activation of a module at a specific layer to its clean state within the corrupted run. This setup enables us to measure the causal contribution of the target component to factual association recall.

Under this setup, we use the metric Indirect Effect (IE) to quantify the degree to which a given component contributes to factual recall. IE is computed as the difference in prediction probability of the correct token $o$ between the restored and non-restored cases~\cite{meng2022locating}:

\begin{equation}
  \label{equation5}
  IE = \mathbb{P}_{*, \text{clean } h_i^{(l)}}[o] - \mathbb{P}_*[o]
\end{equation}

\noindent where $\mathbb{P}_*[o]$ denotes the prediction probability of $o$ in the corrupted run, and $\mathbb{P}_{*, \text{clean } h_i^{(l)}}[o]$ is the corresponding probability when the hidden state $h_i^{(l)}$ is restored to its clean value. In this experiment, IE is used to quantify the causal contribution of the target component, where a higher IE indicates that the activation of the module at the given layer has a greater impact on factual recall.

Finally, the Average Indirect Effect (AIE) aggregates IE across multiple prompts, capturing the degree to which a given component consistently supports factual recall across diverse contexts.\\

\subsection{Severing Effects}
Our second experiment {\bf{\#2}} applies the severing effects to quantitatively analyze the extent to which AIE-based contributions are concentrated in either the Attention or MLP modules. Specifically, similar to the restoration effects, we start from a hidden state in which activations have been restored, but selectively replace only the activation of the target module (e.g., Attention or MLP) with its corrupted counterpart, thereby severing the information flow through that module. If the AIE value significantly decreases when the activation of a specific module at a certain layer is corrupted, it suggests that the corresponding activation plays a critical role in factual association recall.

The restoration and severing effects experiments yield diverging results, making it difficult to consistently interpret the contribution of each module.
To address this discrepancy, we compute the Gini coefficient~\cite{dorfman1979formula} over the AIE distribution to quantify the concentration of contributions across layers.
The Gini coefficient is defined as follows:

\begin{equation}
  \label{equation6}
  G_{AIE}=\frac{\sum_i\sum_j|AIE_i'-AIE_j'|}{2L\sum_iAIE_i'}
\end{equation}

\noindent where $L$ is the number of total layers, and $AIE'$ denotes the normalized, layer-level aggregated AIE values.

We compute the Gini coefficient to identify the layer with concentrated AIE, and then apply the severing effects to evaluate whether the AIE reduction aligns proportionally with the restoration effects.

\begin{figure*}[t]
  \includegraphics[width=\textwidth]{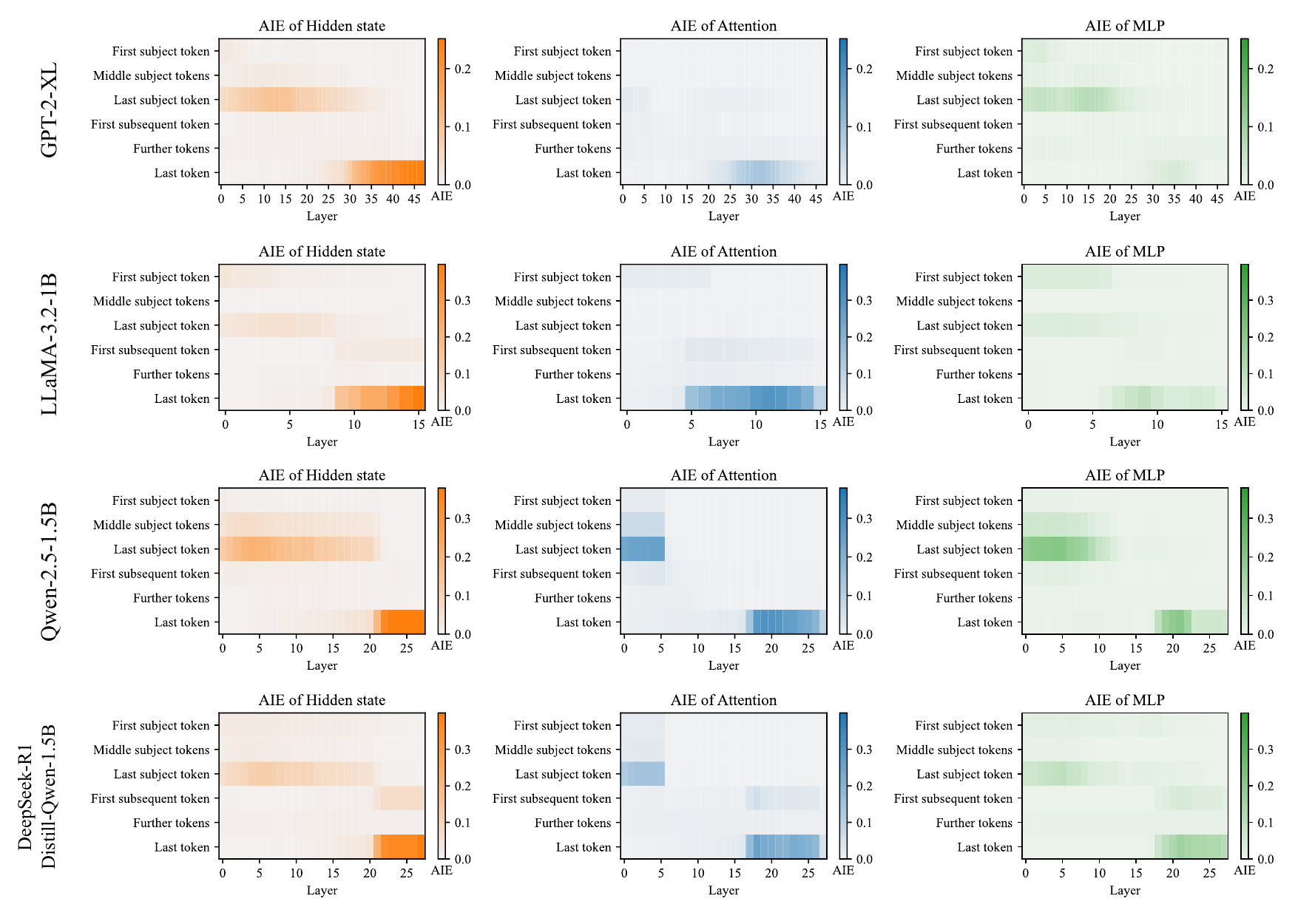}
  \caption{Restoration effects across multiple autoregressive Transformer models.}
  \label{Figure2}
\end{figure*}

\subsection{Factual Prediction}
\label{Section_Factual_Prediction}
Our third experiment {\bf{\#3}} aims to quantitatively evaluate how the Attention and MLP modules inject factual association into the subject representation and functionally contribute to factual prediction. To this end, we execute a {\bf{knockout-run}}, which sequentially applies knockout interventions to each layer \( l = 0, \dots, L - 1 \).  
Specifically, we zero out the updates at the last subject token position in both the Attention and MLP modules across five consecutive layers. The intervention is applied over the range \( \tilde{l} = l, \dots, \min\{l + 4, L - 1\} \), and the model's top-\(k\) output tokens under these intervention conditions are collected. These tokens are subsequently used as the set \( T \), which serves as the basis for factual prediction evaluation.

Prior to evaluation, we construct a candidate object set \( O \) corresponding to each subject. We retrieve relevant paragraphs from Wikipedia using BM25~\cite{bm25, geva2023dissecting}, followed by tokenization and the removal of stopwords and subword fragments. The resulting candidate set consists of non-common tokens that frequently co-occur with the subject and can be considered plausible object expressions.

Language models with a relatively large number of parameters often produce outputs that are semantically accurate but lexically differ from the canonical object tokens. As a result, conventional string match-based metrics significantly underestimate the objects rate. To mitigate this limitation, we adopt a semantic similarity-based evaluation method using Sentence-BERT~\cite{SBERT}.

Therefore, instead of relying on exact string matching, we compute the cosine similarity between the Sentence-BERT embeddings of the generated tokens and the object candidates. Tokens that exceed a predefined similarity threshold \( \tau \) with any element in \( O \) are regarded as semantically valid. This design allows for robust and generalizable evaluation of factual prediction by accounting for lexical variation, improving upon string match-based methods used in prior work~\cite{geva2023dissecting}.

The objects rate is defined as the average proportion of semantically valid tokens among the top-\(k\) outputs for each subject:
\begin{equation}
  \label{equation7}
 ObjectsRate = \frac{|O_\tau|}{|T|} \times 100
\end{equation}

\noindent where \( T \) denotes the set of top-\(k\) tokens (with $k=50$) generated by the model for a given subject, and \( O_\tau \) is defined as:
\begin{equation}
  \label{equation8}
 O_\tau = \left\{ t \in T \;\middle|\; \exists o \in O,\ \mathrm{sim}(t, o) \geq \tau \right\}
\end{equation}
Here, \( O \) is the candidate object set for each subject, and \( \mathrm{sim}(t, o) \) denotes the cosine similarity between the generated token \( t \) and the object candidate \( o \). The similarity threshold is set to \( \tau = 0.7 \)\footnote{Detailed in Appendix~\ref{Appendix_B}.}, and the objects rate is averaged across all subjects to yield a generalized measure of factual prediction performance.

\begin{figure*}[t]
  \includegraphics[width=\textwidth]{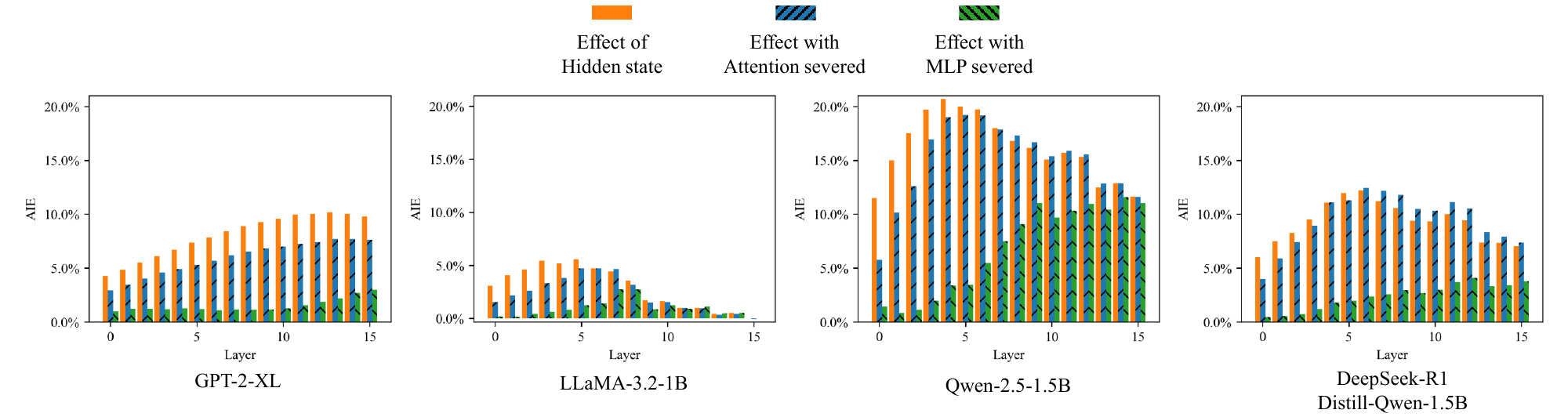}
  \caption{Severing effects across multiple autoregressive Transformer models (restricted to layers 0–15 to verify whether severing early layers significantly reduces AIE, consistent with the concentration observed in the restoration effects experiment).}
  \label{Figure3}
\end{figure*}

\begin{figure}[t]
    \centering
  \includegraphics[width=0.8\linewidth]{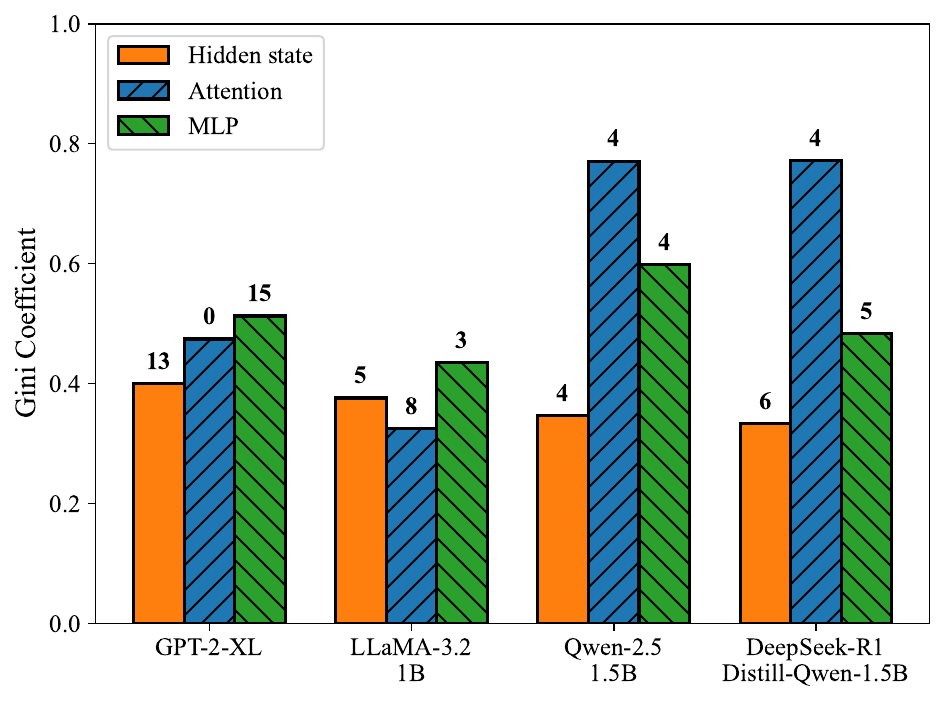}
  \caption{Gini coefficient-based concentration analysis in Attention and MLP modules across autoregressive Transformer models (numbers above bars indicate the layer with the highest concentration).}
  \label{Figure4}
\end{figure}

\begin{figure*}[t]
  \includegraphics[width=\textwidth]{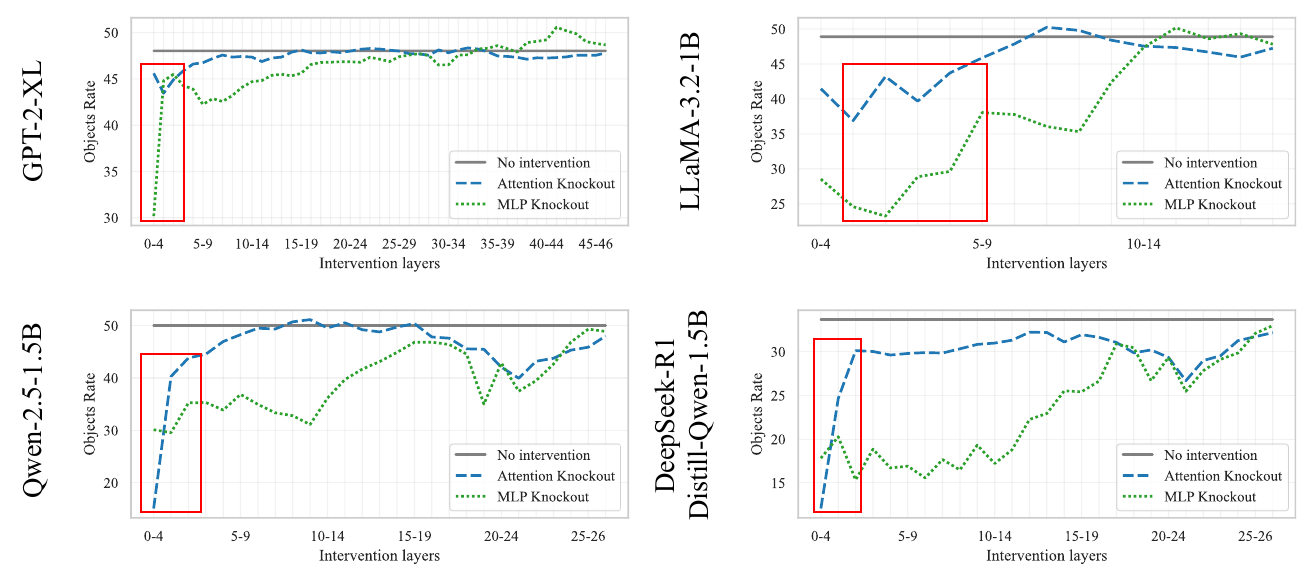}
  \caption{Factual prediction evaluation after knockout of Attention and MLP outputs across autoregressive Transformer models. Red boxes indicate points where blocking either the Attention or MLP module in early layers causes a substantial drop in objects rate, suggesting that the corresponding module plays a critical role in factual association recall.}
  \label{Figure5}
\end{figure*}

\section{Evaluation Results}
\subsection{Restoration Effects Analysis}
\label{Restoration Effect Analysis}
To analyze the components contributing to the recall of factual associations, we use the restoration effects experiment to compute the Average Indirect Effect (AIE). The experiments are conducted on autoregressive Transformer-based models including GPT, LLaMA, Qwen, and DeepSeek, with a particular focus on the last subject token, which plays a crucial role in factual association recall.\\
\indent Figure~\ref{Figure2} shows the results of the restoration effects experiment, visualizing the layer-wise and module-wise AIE distribution across all token positions for each model. In GPT and LLaMA, AIE values are elevated in the early MLP layers, reflecting a canonical pattern of factual recall~\cite{meng2022locating}. By contrast, Qwen and DeepSeek exhibit notably high AIE values in the early layers of the Attention module, suggesting that, unlike GPT and LLaMA, they store factual associations in the early Attention layers.\\

\subsection{Severing Effects Analysis}
\label{Severing Effect Analysis}
To quantitatively analyze whether AIE-based contributions are more concentrated in the Attention or MLP modules, we perform a severing effects experiment in which we remove the outputs of specific layers and measure the resulting changes in AIE. As prior results (Figure~\ref{Figure2}) indicate that the early layers of both modules exhibit relatively high AIE values, we design the experiment to target layers 0 through 15 for each module. 

Figure~\ref{Figure3} shows the results of the severing effects experiment, focusing on layers with high AIE values identified in the previous restoration effects analysis. In GPT and LLaMA, the AIE values are elevated in the early MLP layers, and severing these layers leads to a substantial reduction in AIE. By contrast, in Qwen and DeepSeek, although the early Attention layers exhibit high AIE values, severing these layers results in only a minimal decrease. This indicates that while the Attention module appears to contribute significantly in the restoration effects experiment, its influence is not effectively suppressed in the severing effects setting.

To quantitatively explain this discrepancy, we compute the Gini coefficient of the layer-wise AIE distribution, as shown in Figure~\ref{Figure4}. The results show that in the Qwen and DeepSeek models, the AIE contribution of the Attention module is highly concentrated in the fourth layer. In contrast, in the GPT and LLaMA models, contributions from both the Attention and MLP modules are more evenly distributed.
Based on this analysis, we conduct an additional experiment that selectively severs the layers with the highest concentration of contributions in each module. The results are summarized in Table~\ref{Table2}. Severing the MLP layers leads to a clear drop in AIE, while severing the Attention layers yields only a minimal reduction. This outcome can be attributed to the structural characteristics of the Attention mechanism. In the Attention module, information can propagate through alternative paths even when specific routes are blocked, which mitigates the impact of severing a particular layer~\cite{elhage2021mathematical}. In contrast, the MLP module operates independently on each token, and severing it directly disrupts the flow of information. These structural differences account for the observed disparity in AIE reduction between the two modules in the severing effects experiment~\cite{geva_mlp_key_value}.


\begin{table}[t]
\centering
\begin{tabular}{ccc}
\Xhline{1.2pt}
\textbf{Model} & \makecell[c]{\textbf{Attention}\\(Drop rate)} & \makecell[c]{\textbf{MLP}\\(Drop rate)} \\
\Xhline{0.8pt}
GPT-2-XL & 31.32\% & 69.52\% \\
\Xhline{0.2pt}
LLaMA-3.2-1B & 10.82\% & 89.40\% \\
\Xhline{0.2pt}
Qwen-2.5-1.5B & 8.25\% & 83.55\% \\
\Xhline{0.2pt}
\begin{tabular}[c]{@{}c@{}}DeepSeek-R1\\Distill-Qwen-1.5B\end{tabular} & -0.14\% & 83.60\% \\
\Xhline{1.2pt}
\end{tabular}
\caption{Effect of severing highly concentrated layers selected by the Gini coefficient on AIE. A larger drop rate indicates greater layer influence (for Qwen and DeepSeek, severing Attention layers with high Gini scores results in only a small drop).}
\label{Table2}
\end{table}

\subsection{Factual Prediction Analysis}
\label{Factual Prediction Analysis}
To more precisely evaluate the contribution of the Attention module, we conduct a factual prediction experiment using a knockout approach that completely blocks the output of the target module.

Figure~\ref{Figure5} shows the change in objects rate when a knockout is applied to each layer of the Attention and MLP modules. In GPT and LLaMA, blocking the early MLP layers leads to a substantial drop in objects rate, whereas in Qwen and DeepSeek, blocking the early Attention layers also results in a substantial drop. These results suggest that GPT and LLaMA primarily store factual associations in the early layers of the MLP module, while Qwen and Qwen-based DeepSeek store a substantial portion of such associations in the early layers of the Attention module, which plays a critical role in the recall of factual associations. These findings indicate that the factual prediction experiment provides a more appropriate and reliable means of assessing the contribution of the Attention module than the severing effects experiment.

\section{Practical Implications}
\label{s:discuss}
Our study highlights important differences in how language models store and recall facts, depending on their architecture, which have important practical implications.
For example, in Qwen-based architectures, where factual recall is concentrated in early Attention layers, knowledge editing methods~\cite{meng2022locating,meng2022mass,li2024pmet,fang2024alphaedit} should adapt accordingly: targeting Attention modules more than MLPs. However, the underlying causes of these architectural divergences remain unclear. To further probe this issue, we present and formulate two hypotheses related to subject tokenization and architectural design factors, and conduct exploratory analyses, the results of which are presented in Appendix~\ref{Appendix_D}.\\
\indent Similarly, interpretability tools~\cite{clark-etal-2019-bert, belinkov2022probing} and attribution analyses~\cite{vig-2019-multiscale} should focus on these components to better trace factual reasoning.
These insights can also guide architecture-aware model compression~\cite{sanh2020movement}, where preserving key Attention layers may help retain factual knowledge.


\section{Related Work}
Language models go beyond predicting sequences based solely on frequently occurring word patterns, as they internally store and utilize factual associations~\cite{petroni2019language,jiang2020can, roberts-etal-2020-much}.
Early research on factual association recall focuses on probing methods, which train classifiers on frozen representations to evaluate whether specific properties are embedded within the internal representations of language models~\cite{ettinger2016probing,adi2016fine,belinkov2017neural,hupkes2018visualisation,conneau-etal-2018-cram,elazar2021measuring}. However, such approaches are limited in explaining how these properties functionally influence model predictions~\cite{belinkov2022probing}.

To address these limitations, recent work turns to causal analysis, which estimates the functional contributions of internal components by applying counterfactual interventions~\cite{vig2020investigating,pearl2022direct}. A prominent method that applies this concept to hidden representation analysis is causal tracing, which enables the layer-level and module-level attribution of factual predictions~\cite{meng2022locating}.
Recent work actively explores causal tracing, which now serves as a central analytical framework for model interpretability~\cite{dai-etal-2022-knowledge,mohebbi2023quantifying,hase2023does,geva2023dissecting,dar-etal-2023-analyzing} and knowledge editing~\cite{meng2022locating, meng2022mass, li2024pmet, fang2024alphaedit}.

\section{Conclusion}
We conducted a comprehensive evaluation of factual association recall to examine whether prior findings based on GPT models generalize across a broader range of autoregressive Transformer architectures.
To this end, we designed our evaluation methods to highlight structural differences in factual recall behavior across models.
Our results reveal that Qwen-based models behave differently from previously reported patterns: factual recall is more strongly concentrated in the Attention modules, rather than the MLPs.
Our findings have important implications for the deployment of Transformer models in real-world applications. 
In particular, understanding which layers and modules are responsible for storing factual knowledge is essential for improving performance in knowledge-intensive tasks, enabling more precise model editing, and supporting interpretability.

\section*{Limitations}
While we analyze where factual associations are stored across a range of autoregressive Transformer models, including GPT, LLaMA, Qwen, and DeepSeek, the internal mechanisms by which these associations are recalled during inference remain insufficiently understood. Furthermore, despite observing model-specific recall patterns, how factual associations are dynamically accessed across different architectures remains unclear. In addition, although our results indicate that the storage locations of factual associations may vary depending on the model architecture, the interpretation of such structural differences is limited, warranting further investigation. Addressing these limitations is crucial for applications such as knowledge editing across existing and emerging Transformer-based architectures.

\section*{Acknowledgments}
We thank the anonymous reviewers and the area chair for their constructive comments. This work was supported by Institute of Information \& communications Technology Planning \& Evaluation (IITP) grant funded by the Korea government (MSIT) (No.RS-2024-00398353, Development of Countermeasure Technologies for Generative AI Security Threats). 

\bibliography{custom}


\appendix

\section{Additional Experimental Analysis}
\label{Appendix_A}
We extend our experiments to autoregressive Transformer models with varying parameter sizes, as listed in Table~\ref{Table3}, and conduct additional evaluations accordingly.

\subsection{Restoration Effects Analysis}
\begin{figure*}[t]
	\includegraphics[width=\textwidth]{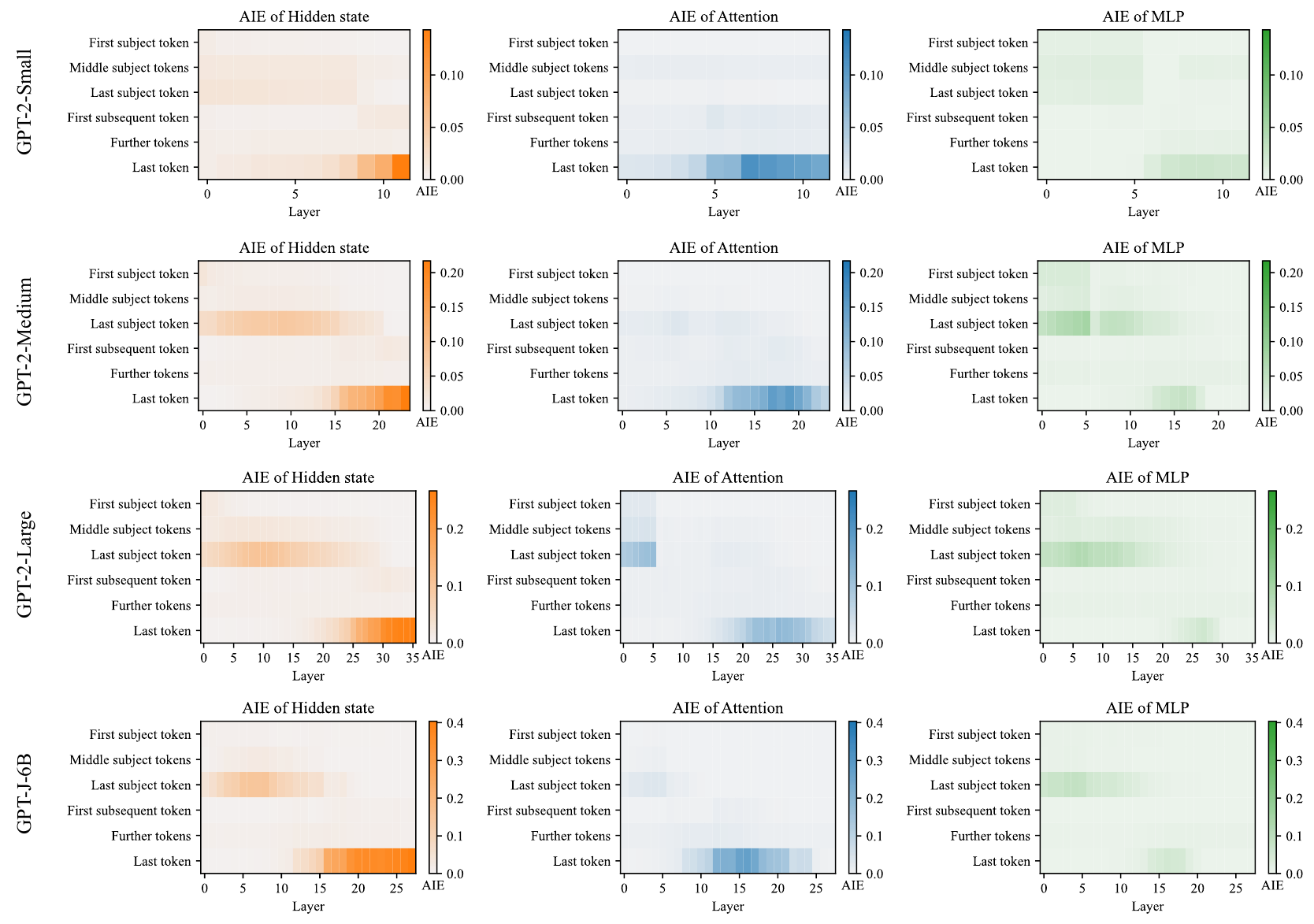}
	\caption{Restoration effects in GPT-family autoregressive Transformer models with varying parameter scales.}
	\label{appendix_figure_restoration_gpt}
\end{figure*}

\begin{figure*}[t]
	\includegraphics[width=\textwidth]{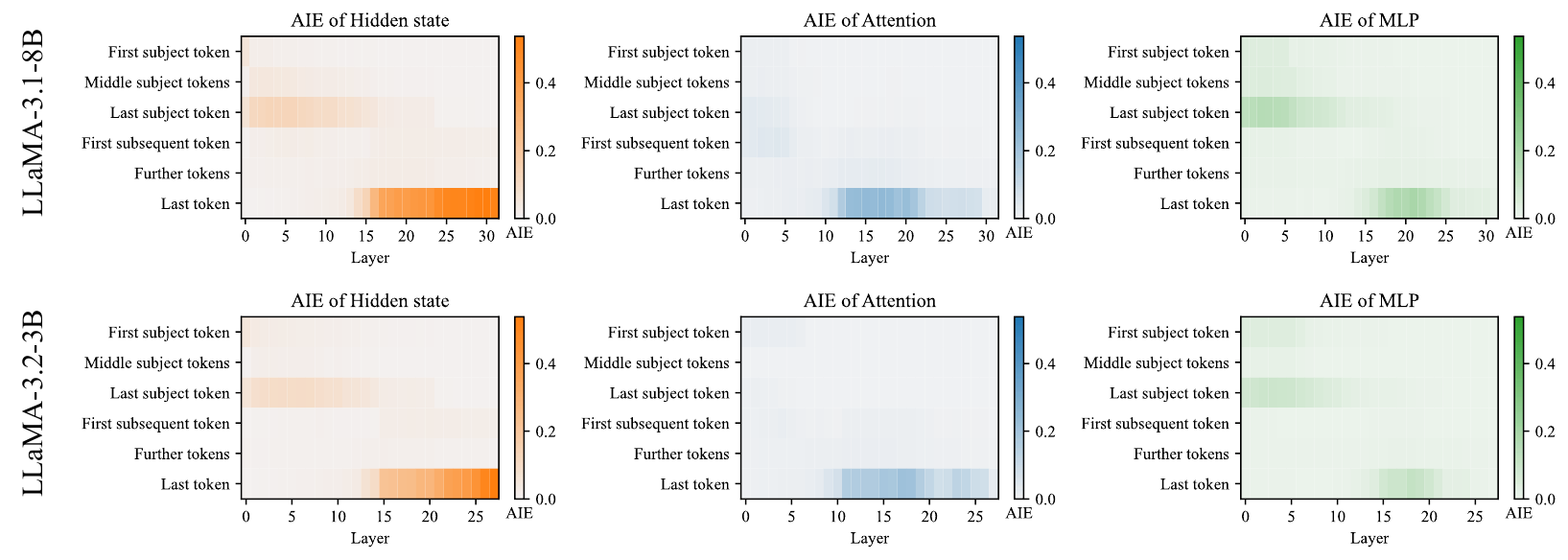}
	\caption{Restoration effects in LLaMA-family autoregressive Transformer models with varying parameter scales.}
	\label{appendix_figure_restoration_llama}
\end{figure*}

\begin{figure*}[t]
	\includegraphics[width=\textwidth]{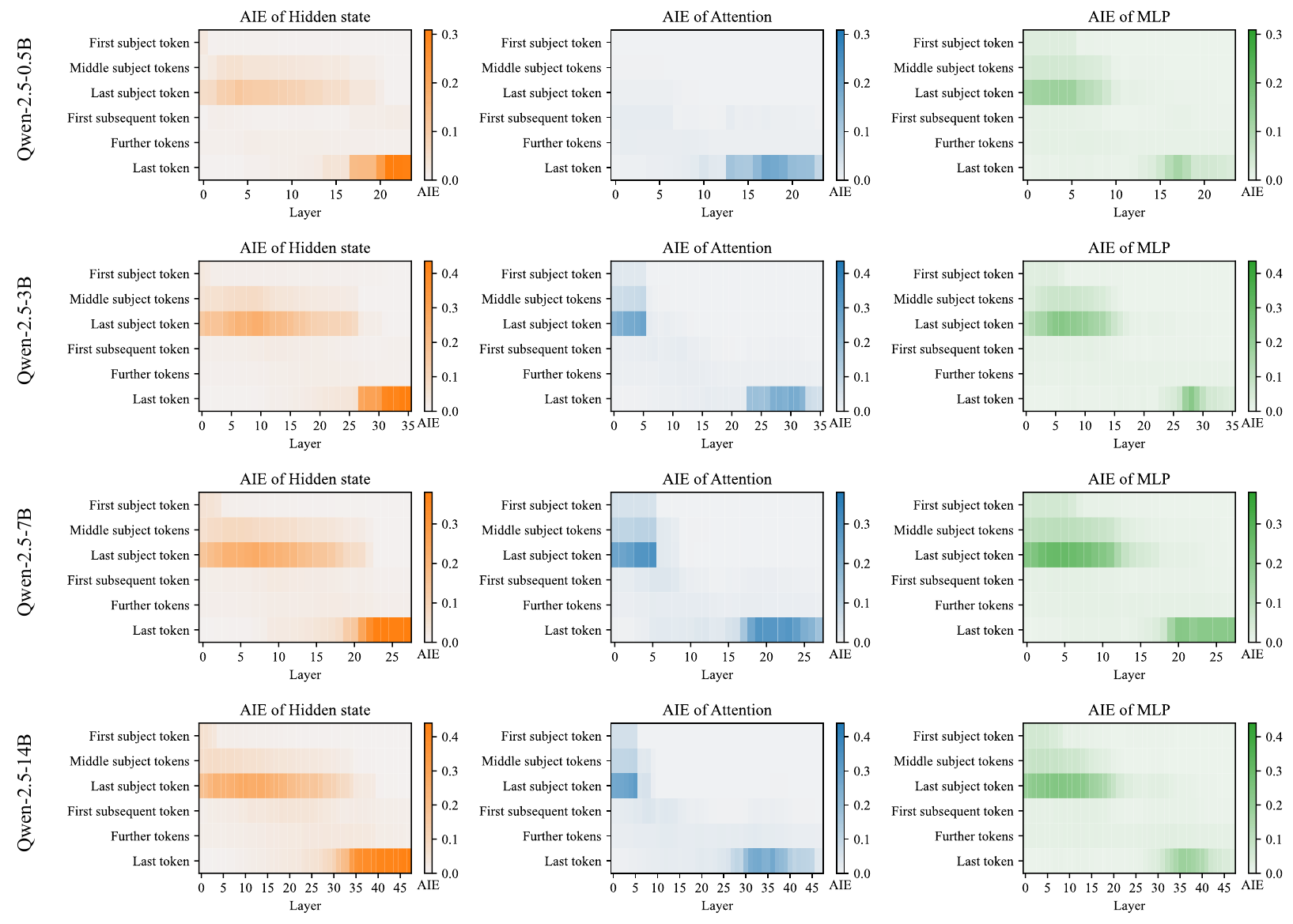}
	\caption{Restoration effects in Qwen-family autoregressive Transformer models with varying parameter scales.}
	\label{appendix_figure_restoration_qwen}
\end{figure*}

\begin{figure*}[t]
	\includegraphics[width=\textwidth]{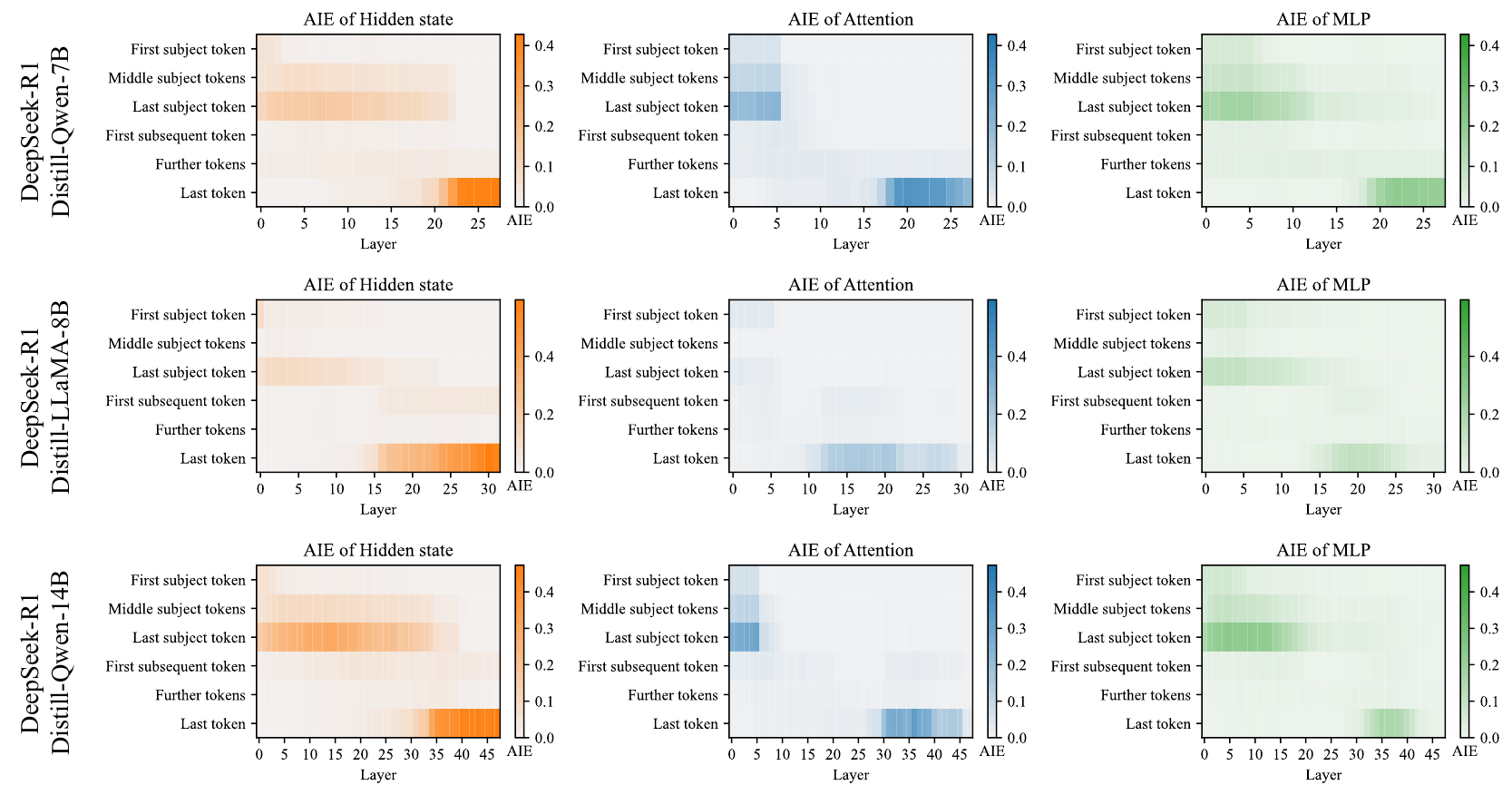}
	\caption{Restoration effects in DeepSeek-family autoregressive Transformer models with varying parameter scales.}
    \label{appendix_figure_restoration_deepseek}
\end{figure*}

We extend the restoration effects experiment by conducting additional analyses on autoregressive Transformer models with varying parameter scales.\\
\indent We show the restoration effects results separately for each model family. Figure~\ref{appendix_figure_restoration_gpt} shows the results for the GPT family, and Figures \ref{appendix_figure_restoration_llama}, \ref{appendix_figure_restoration_qwen}, and \ref{appendix_figure_restoration_deepseek} show the results for the LLaMA, Qwen, and DeepSeek families, respectively.\\
\indent First, the analysis of GPT-family models reveals that the MLP modules in the early layers at the last subject token play a primary role in storing factual associations. Notably, GPT-2 Large exhibits a distinct pattern, where the Attention modules in the early layers also make a significant contribution to recalling factual associations.\\
\indent Second, the analysis of LLaMA-family models reveals that, similar to the GPT family, the MLP modules in the early layers at the last subject token play a major role in storing factual associations.\\
\indent Third, the analysis of Qwen-family models reveals that both MLP and Attention modules contribute to the storage of factual associations primarily in the early layers at the last subject token. This suggests that Qwen models leverage multiple modules within the early layers jointly in the process of storing factual associations.\\
\indent Finally, the analysis of DeepSeek-family models reveals that the pattern of factual association storage varies depending on the base model used for distillation. Qwen-based models store factual associations in the early layers at the last subject token through both Attention and MLP modules, whereas LLaMA-based models show no clear contribution from the Attention modules at the same position. Furthermore, the distillation models exhibit factual association recall structures similar to their base models and display generally consistent patterns.\\
\indent These results suggest that the location of factual association storage may vary depending on structural differences in model architecture.\\
\indent Prior work, such as ROME~\cite{meng2022locating}, suggests that in models with a small number of layers, factual associations may be stored in the early layers of the attention module at the position of the last subject token. However, our experimental results show that clear activation is not observed in the attention module at the last subject token position in models with a small number of layers, such as GPT-Small, GPT-Medium, and LLaMA-3.2–3B. These findings suggest that the contribution of the attention module to factual association storage is not determined solely by the number of layers, but may also be influenced by other architectural factors.

\begin{table}[t]
\centering
\begin{tabular}{ccc}
\Xhline{1.2pt}
\textbf{Model} & \textbf{\#Layers} & \textbf{\#Parameters} \\
\Xhline{0.8pt}
GPT-2-Small & 12 & 0.124B \\
\Xhline{0.2pt}
GPT-2-Medium & 24 & 0.335B \\
\Xhline{0.2pt}
GPT-2-Large & 36 & 0.774B \\
\Xhline{0.2pt}
GPT-J-6B & 28 & 6B \\
\Xhline{0.2pt}
LLaMA-3.1-8B & 32 & 8B \\
\Xhline{0.2pt}
LLaMA-3.2-3B & 28 & 3B \\
\Xhline{0.2pt}
Qwen-2.5-0.5B & 24 & 0.5B \\
\Xhline{0.2pt}
Qwen-2.5-3B & 36 & 3B \\
\Xhline{0.2pt}
Qwen-2.5-7B & 28 & 7B \\
\Xhline{0.2pt}
Qwen-2.5-14B & 48 & 14B \\
\Xhline{0.2pt}
\begin{tabular}[c]{@{}c@{}}DeepSeek-R1\\Distill-Qwen-7B\end{tabular} & 28 & 7B \\
\Xhline{0.2pt}
\begin{tabular}[c]{@{}c@{}}DeepSeek-R1\\Distill-LLaMA-8B\end{tabular} & 32 & 8B \\
\Xhline{0.2pt}
\begin{tabular}[c]{@{}c@{}}DeepSeek-R1\\Distill-Qwen-14B\end{tabular} & 48 & 14B \\
\Xhline{1.2pt}
\end{tabular}
\caption{Architectural configurations of autoregressive Transformer models with varying parameter scales (B = billion parameters).}
\label{Table3}
\end{table}

\subsection{Severing Effects Analysis}
\begin{figure*}[t]
	\includegraphics[width=\textwidth]{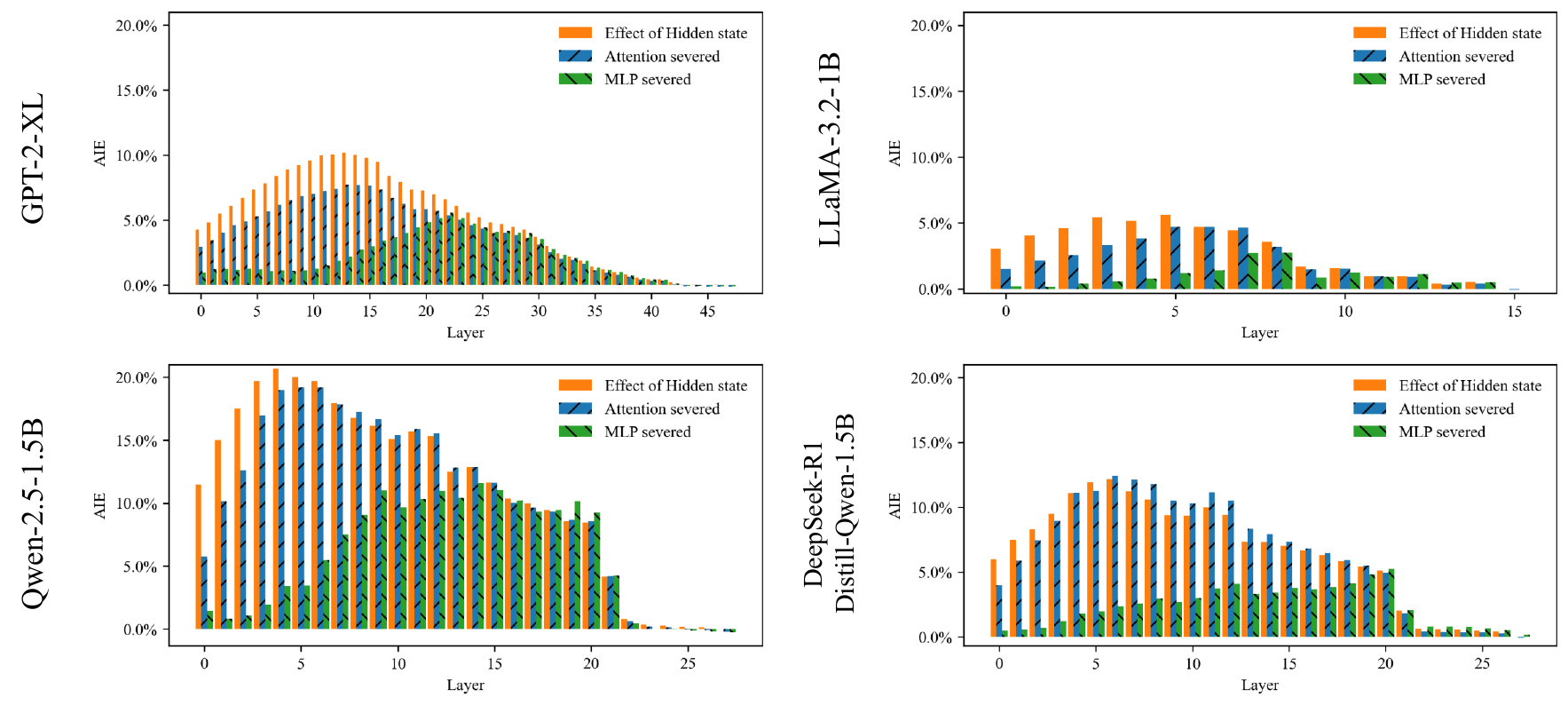}
	\caption{Severing effects across multiple autoregressive Transformer models (all layers).}
	\label{appendix_figure_modified_main_result}
\end{figure*}

\begin{figure*}[t]
	\includegraphics[width=\textwidth]{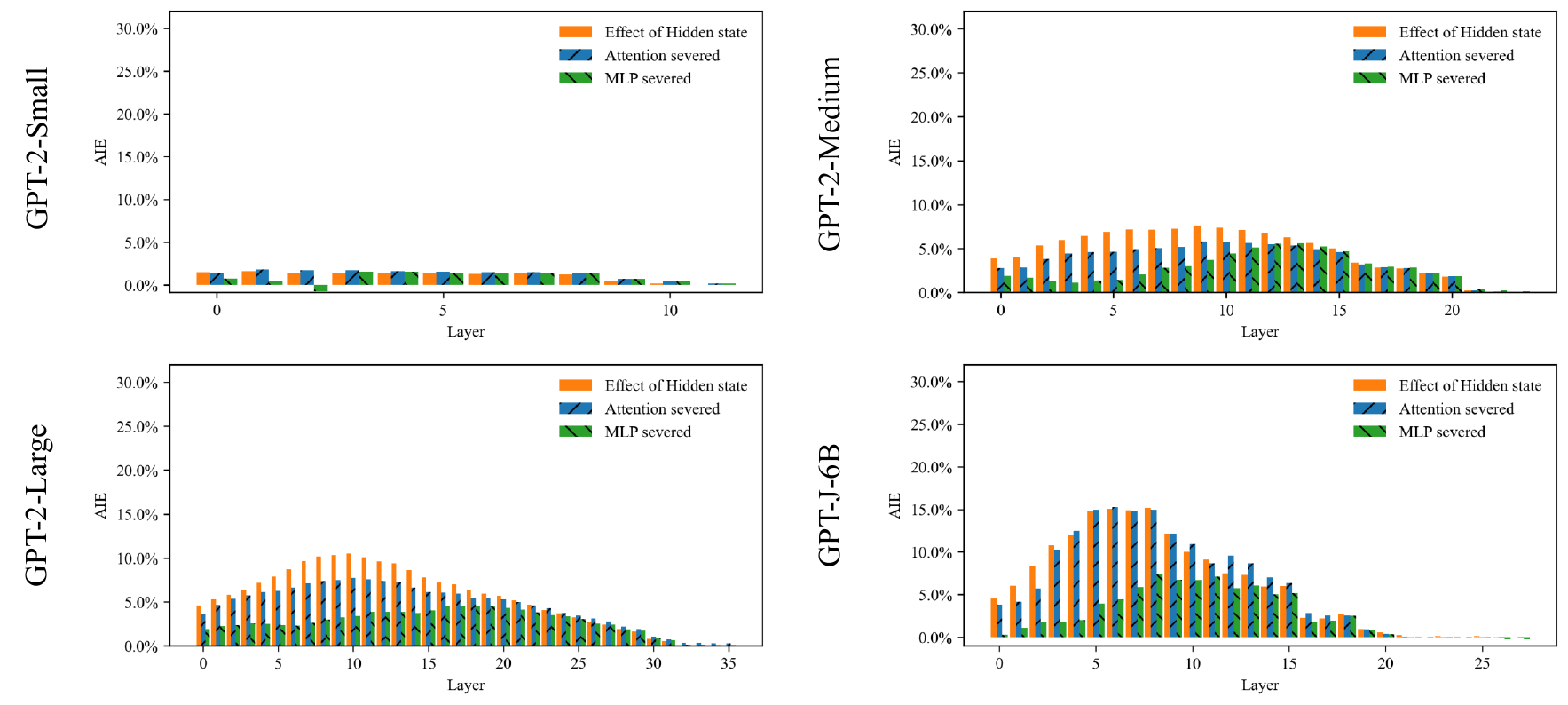}
	\caption{Severing effects across GPT-family autoregressive Transformer models at varying parameter scales.}
	\label{appendix_figure_modified_gpt}
\end{figure*}

\begin{figure*}[t]
	\includegraphics[width=\textwidth]{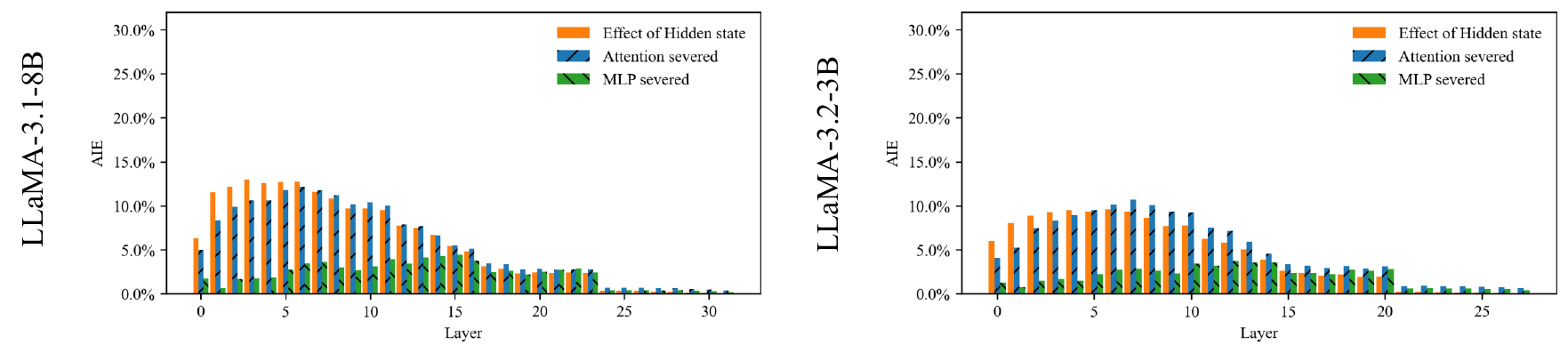}
	\caption{Severing effects across LLaMA-family autoregressive Transformer models at varying parameter scales.}
	\label{appendix_figure_modified_llama}
\end{figure*}

\begin{figure*}[t]
	\includegraphics[width=\textwidth]{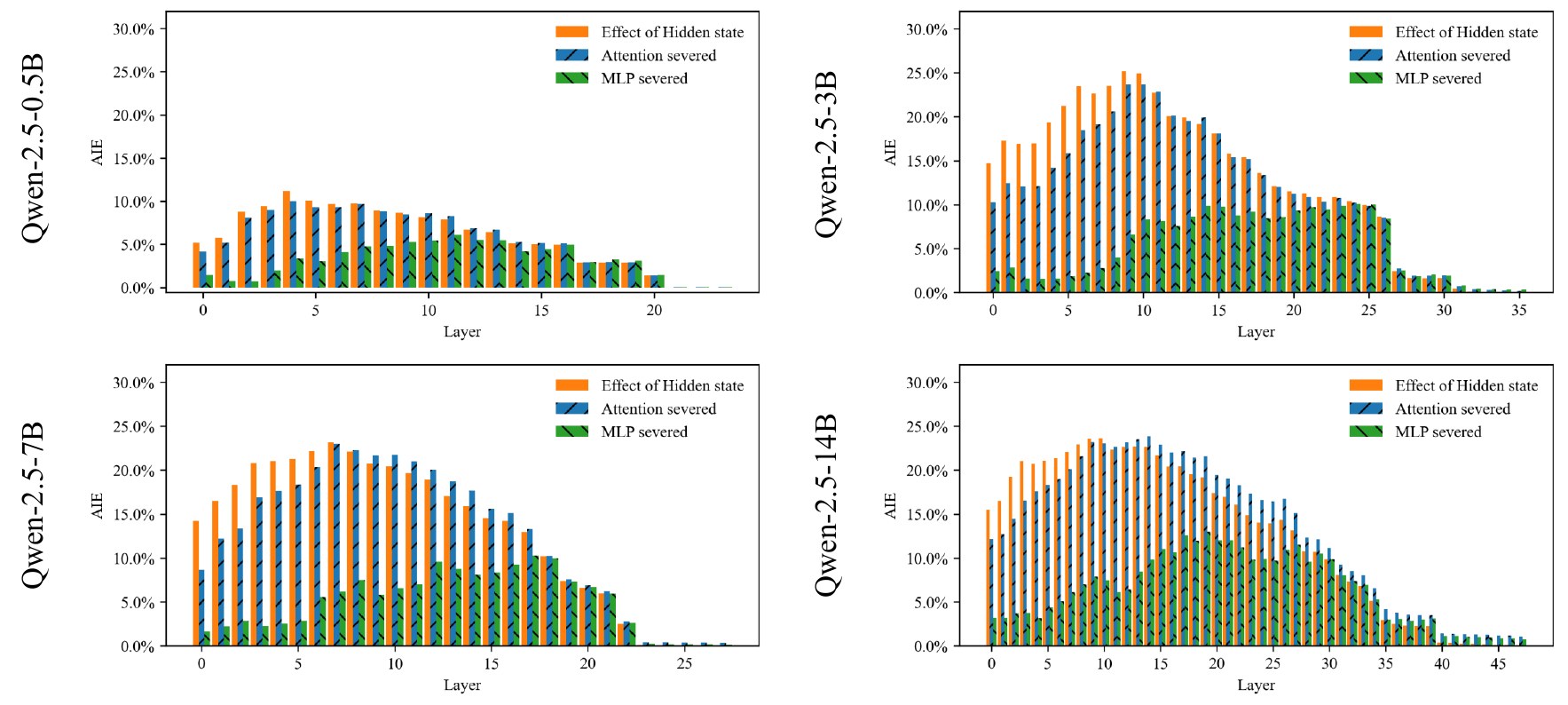}
	\caption{Severing effects across Qwen-family autoregressive Transformer models at varying parameter scales.}
	\label{appendix_figure_modified_qwen}
\end{figure*}

\begin{figure*}[t]
	\includegraphics[width=\textwidth]{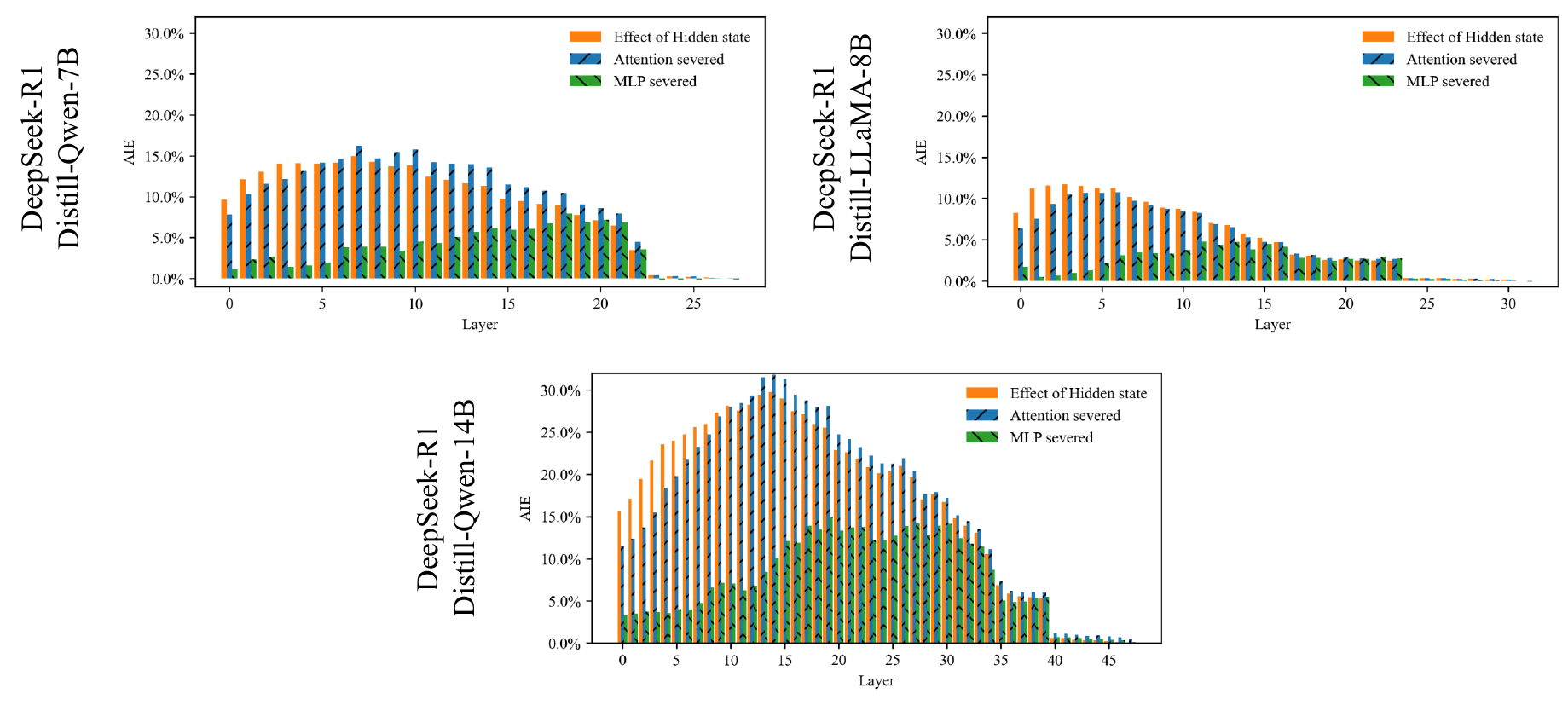}
	\caption{Severing effects across DeepSeek-family autoregressive Transformer models at varying parameter scales.}
	\label{appendix_figure_modified_deepseek}
\end{figure*}

\begin{figure*}[t]
	\includegraphics[width=\textwidth]{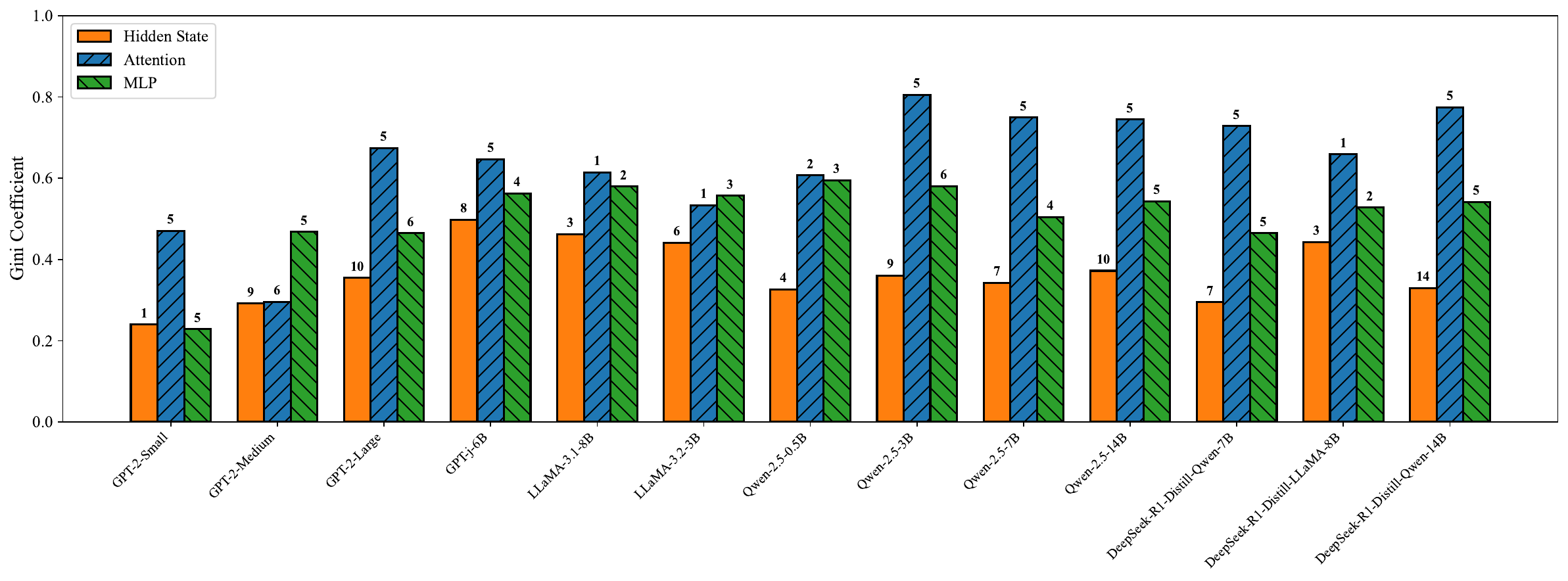}
	\caption{Gini coefficient-based concentration analysis of causal effects in Attention and MLP modules across autoregressive Transformer models with varying parameter scales (numbers above bars indicate the layer with the highest concentration).}
	\label{all_gini_coefficient}
\end{figure*}

We extend the original severing effects experiment, which was limited to layers 0-15, to cover all layers, and perform additional analyses on autoregressive Transformer models with varying parameter scales.\\
\indent Figure~\ref{appendix_figure_modified_main_result} shows the results of the extended severing effects experiments conducted across the full layer range. Figures~\ref{appendix_figure_modified_gpt}, \ref{appendix_figure_modified_llama}, \ref{appendix_figure_modified_qwen}, and \ref{appendix_figure_modified_deepseek} show the outcomes for the GPT, LLaMA, Qwen, and DeepSeek families, respectively.
\indent The analysis of all model families --- including GPT, LLaMA, Qwen, and DeepSeek --- reveals that severing the MLP modules in the early layers at the last subject token leads to a substantial drop in AIE, indicating their critical role in factual association recall. Notably, while Qwen and DeepSeek models exhibit high AIE values for the early Attention modules at the last subject token in the severing effects experiments, the severing effects experiments show that severing these Attention modules does not result in as large a decrease in AIE as observed for the MLP modules.\\
\indent Based on these results, we extend the previous Gini coefficient-based concentration experiments by conducting additional analyses across a range of autoregressive Transformer models to quantitatively examine the distribution characteristics of AIE within the Attention and MLP modules. Figure~\ref{all_gini_coefficient} shows the results of the Gini coefficient-based concentration analysis. Most GPT-family and LLaMA-family models, except for GPT-2-Small and GPT-2-Large, exhibit a relatively uniform distribution of AIE across the Attention and MLP modules. In contrast, Qwen-family and DeepSeek-family models show a concentrated distribution of AIE within specific layers of the Attention module, indicating that factual associations tend to be stored in a more concentrated manner within certain layers of the Attention module.

\subsection{Factual Prediction Analysis}
\begin{figure*}[t]
	\includegraphics[width=\textwidth]{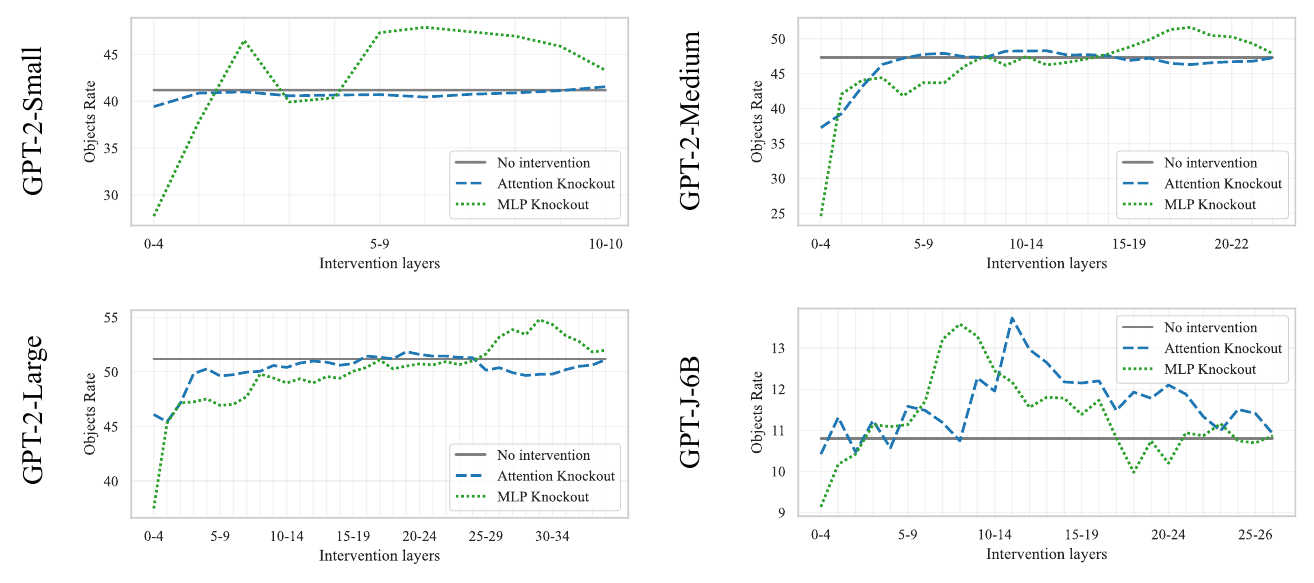}
	\caption{Factual prediction evaluation after knockout of Attention and MLP outputs across GPT-family autoregressive Transformer models with varying parameter scales.}
	\label{appendix_figure_knockout_gpt}
\end{figure*}

\begin{figure*}[t]
	\includegraphics[width=\textwidth]{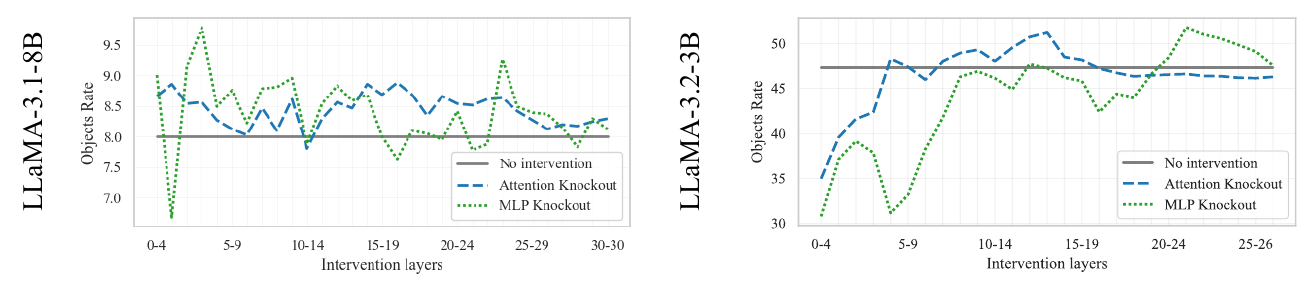}
	\caption{Factual prediction evaluation after knockout of Attention and MLP outputs across LLaMA-family autoregressive Transformer models with varying parameter scales.}
	\label{appendix_figure_knockout_llama}
\end{figure*}

\begin{figure*}[t]
	\includegraphics[width=\textwidth]{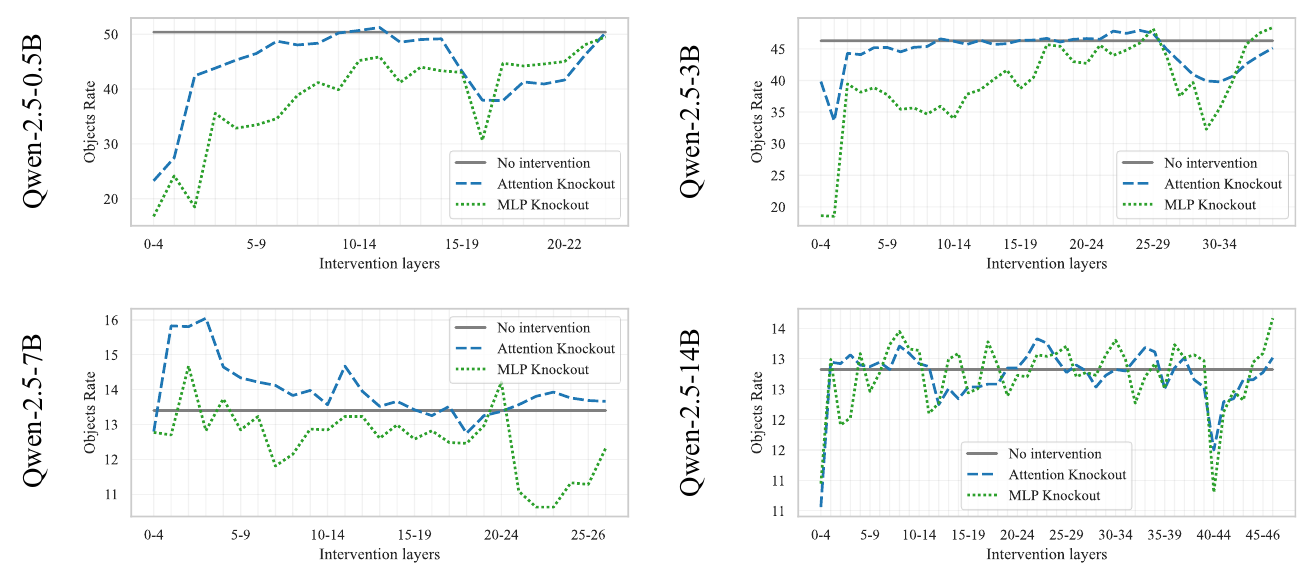}
	\caption{Factual prediction evaluation after knockout of Attention and MLP outputs across Qwen-family autoregressive Transformer models with varying parameter scales.}
	\label{appendix_figure_knockout_qwen}
\end{figure*}

\begin{figure*}[t]
	\includegraphics[width=\textwidth]{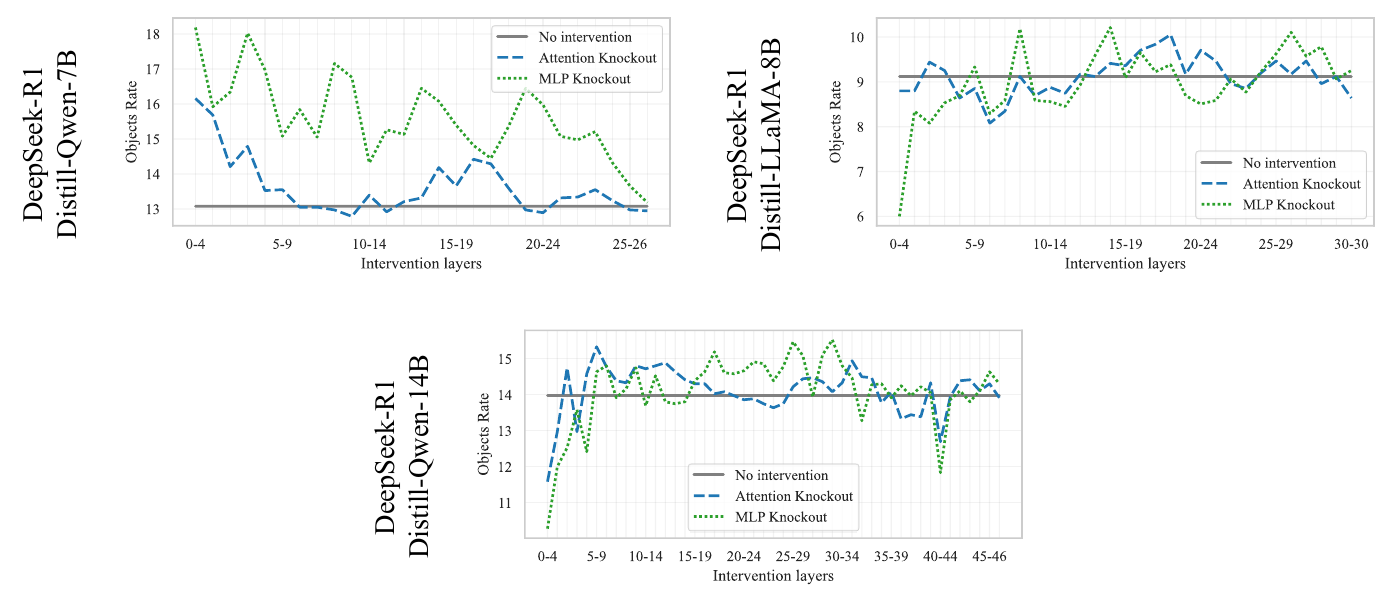}
	\caption{Factual prediction evaluation after knockout of Attention and MLP outputs across DeepSeek-family autoregressive Transformer models with varying parameter scales.}
	\label{appendix_figure_knockout_deepseek}
\end{figure*}

We extend the previous factual prediction experiments by conducting additional evaluations on a variety of autoregressive Transformer models.\\
\indent The factual prediction results are presented separately for each model family. Figure~\ref{appendix_figure_knockout_gpt} presents the results for the GPT family, and Figures~\ref{appendix_figure_knockout_llama}, \ref{appendix_figure_knockout_qwen}, and \ref{appendix_figure_knockout_deepseek} correspond to the LLaMA, Qwen, and DeepSeek families, respectively.\\
\indent Most GPT-family models exhibit a substantial drop in objects rate when the early layers of the MLP module are blocked, suggesting that these layers are key locations where factual associations are recalled.\\
\indent Within the LLaMA family, LLaMA 3.1-8B shows a similar pattern to GPT models, with a significant objects rate drop when early MLP layers are blocked, indicating that factual associations are primarily recalled in that region. In contrast, LLaMA 3.2-3B shows a clear drop in objects rate when early Attention layers are blocked, implying that factual recall primarily occurs in the early Attention module.\\
\indent Most models in the Qwen family also demonstrate that early Attention layers play a central role in factual association recall, as evidenced by marked decreases in objects rate when these layers are blocked.\\
\indent For the DeepSeek family, DeepSeek-R1-Distill-LLaMA-8B exhibits factual recall concentrated in the early MLP layers, while DeepSeek-R1-Distill-Qwen-14B shows that both early Attention and MLP layers contribute to factual recall.
Meanwhile, DeepSeek-R1-Distill-Qwen-7B produces inconsistent results, indicating the need for further investigation in future work.

\section{Semantic Similarity Threshold}
\label{Appendix_B}
This section explains the rationale behind the semantic similarity threshold of 0.7 used in the factual prediction experiment.\\
\indent Language models with a relatively large number of parameters exhibit expressive capabilities. As a result, evaluating their outputs solely based on surface-level string matching is insufficient to capture their semantic appropriateness. To address this limitation, this study adopts a quantitative evaluation method based on semantic similarity.\\
\indent For computing semantic similarity, this study employs all-MiniLM-L6-v2, a pre-trained language model based on Sentence-BERT~\cite{SBERT}. Each word pair is encoded into embeddings, and cosine similarity is used to quantify their semantic closeness. Pairs with a similarity score of 0.7 or higher are considered semantically similar. This threshold is selected as a conservative and empirically grounded criterion to ensure a clearer separation between semantically similar and dissimilar cases, thereby enhancing consistency and precision in evaluation. Although lower similarity scores may sometimes correspond to semantically related expressions, a threshold of 0.7 is adopted to maintain reliability and interpretability.\\
\indent To support the validity of this threshold, Table~\ref{Table4} presents examples of sentence pairs spanning a range of similarity scores. These examples demonstrate that semantic similarity–based evaluation enables more fine-grained and accurate assessment than methods relying solely on surface-level string matching.

\begin{table}[t]
\centering
\begin{tabular}{ccc}
\Xhline{1.2pt}
\textbf{Word A} & \textbf{Word B} & \textbf{Similarity Score} \\
\Xhline{0.8pt}
table     & sadness     & 0.09 \\
law       & noodles     & 0.17 \\
rocket    & freedom     & 0.24 \\
chair     & depression  & 0.32 \\
urinary   & water       & 0.40 \\
task      & project     & 0.56 \\
publicity & advertising & 0.62 \\
student   & school      & 0.65 \\
movie     & cinema      & 0.68 \\
\textbf{combat}    & \textbf{fight}       & \textbf{0.71} \\
\textbf{computer}  & \textbf{laptop}      & \textbf{0.71} \\
\textbf{art}       & \textbf{painting}    & \textbf{0.72} \\
\textbf{middle}    & \textbf{mid}         & \textbf{0.75} \\
\textbf{city}      & \textbf{urban}       & \textbf{0.86} \\
\textbf{purchase}  & \textbf{buy}         & \textbf{0.87} \\
\textbf{bike}      & \textbf{bicycle}     & \textbf{0.92} \\
\textbf{sofa}      & \textbf{sofa}        & \textbf{1.00} \\
\Xhline{1.2pt}
\end{tabular}
\caption{Examples of semantic similarity scores for word pairs measured by Sentence-BERT.}
\label{Table4}
\end{table}

\section{Implementation Details}
\label{Appendix_C}
All language model families used in our experiments—including GPT, LLaMA, Qwen, and DeepSeek—are loaded via the Hugging Face Transformers\footnote{\url{https://huggingface.co/}} library, which provides a unified interface for model and tokenizer handling.

During preprocessing as described in Section~\ref{Section_Factual_Prediction}, we use NLTK\footnote{\url{https://www.nltk.org/}} to remove stopwords from prompts.
For candidate selection, we apply BM25 ranking using the rank\_bm25\footnote{\url{https://github.com/dorianbrown/rank_bm25}} python package.

\citet{meng2022locating} is released under the MIT License, and \citet{geva2023dissecting} under the Apache License 2.0, both of which we build upon and modify to support our evaluation framework and to extend their functionalities for a broader set of autoregressive Transformer architectures.

\begin{figure*}[t]
	\includegraphics[width=\textwidth]{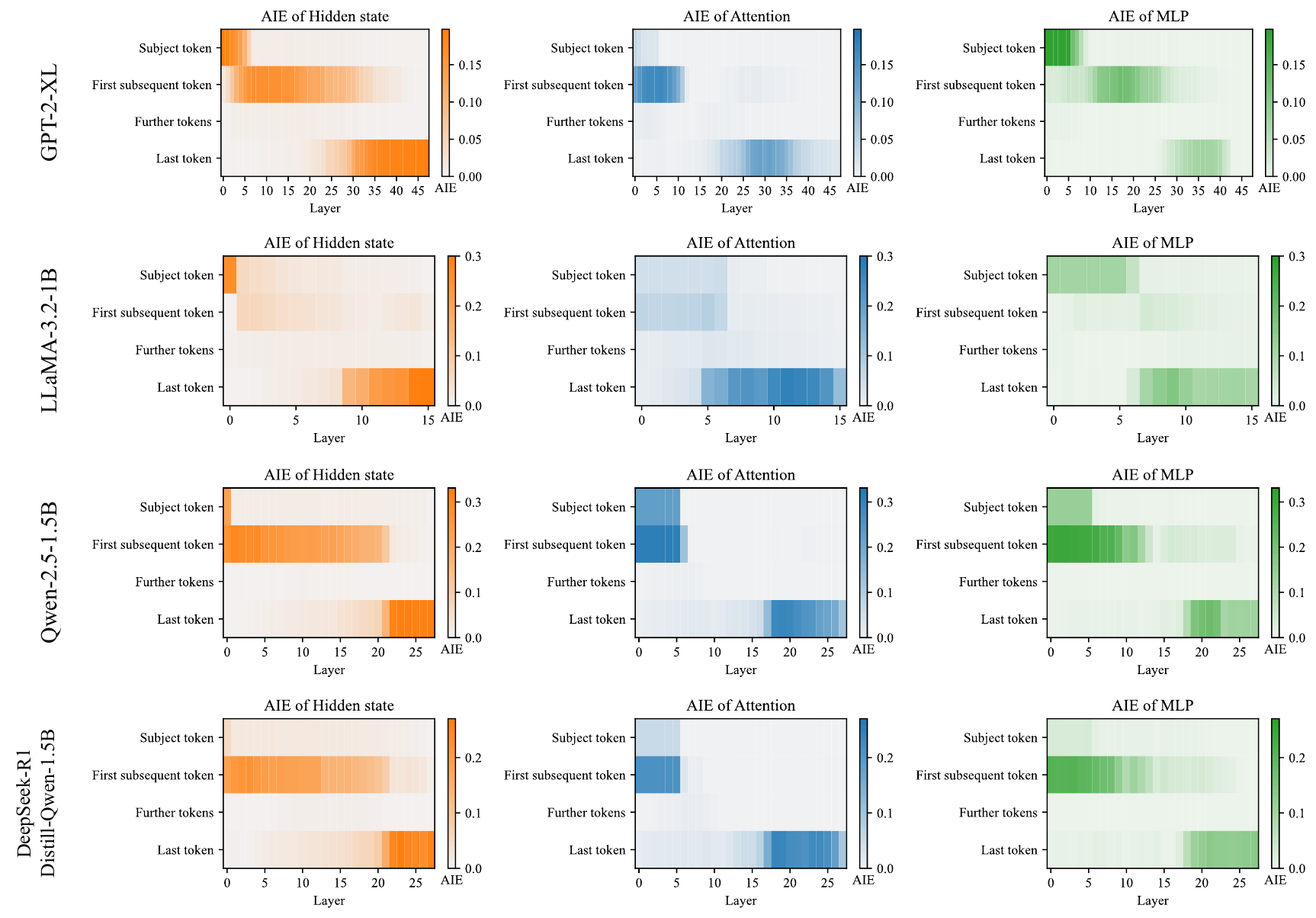}
	\caption{Restoration effects across multiple autoregressive Transformer models under the single-token subject setting.}
	\label{appendix_figure_single_token}
\end{figure*}


\begin{table*}[t]
\centering
\makebox[\textwidth][c]{
\begin{tabular}{ccccccc}
\Xhline{1.2pt}
\textbf{Model} & 
\begin{tabular}[c]{@{}c@{}}\textbf{MLP}\\\textbf{Hidden Size}\end{tabular} & 
\begin{tabular}[c]{@{}c@{}}\textbf{Total}\\\textbf{Attention Heads}\\\textbf{(head × layer)}\end{tabular} & 
\begin{tabular}[c]{@{}c@{}}\textbf{Attention}\\\textbf{Mechanism}\end{tabular} & 
\begin{tabular}[c]{@{}c@{}}\textbf{Recall}\\\textbf{Concentration}\end{tabular} &
\textbf{Tokenizer} &
\begin{tabular}[c]{@{}c@{}}\textbf{Voca}\\\textbf{Size}\end{tabular} \\
\Xhline{0.8pt}
GPT-2 XL & 6,400 & 25 × 48 = 1,200 & MHA & Early MLP & BPE & 50,257 \\
\Xhline{0.2pt}
LLaMA-3.2-1B & 8,192 & 32 × 16 = 512 & GQA & Early MLP & BPE & 128,256 \\
\Xhline{0.2pt}
LLaMA-3.1-8B & 14,336 & 32 × 32 = 1,024 & GQA & Early MLP & BPE & 128,256 \\
\Xhline{0.2pt}
Qwen-2.5-1.5B & 8,960 & 12 × 28 = 336 & GQA & Early Attention & BPE & 151,936 \\
\Xhline{0.2pt}
\begin{tabular}[c]{@{}c@{}}DeepSeek-R1\\Distill-Qwen-1.5B\end{tabular} & 8,960 & 12 × 28 = 336 & GQA & Early Attention & BPE & 151,936 \\
\Xhline{0.2pt}
Qwen-2.5-14B & 13,824 & 40 × 48 = 1,920 & GQA & Early Attention & BPE & 152,064 \\
\Xhline{0.2pt}
\begin{tabular}[c]{@{}c@{}}DeepSeek-R1\\Distill-Qwen-14B\end{tabular} & 13,824 & 40 × 48 = 1,920 & GQA & Early Attention & BPE & 152,064 \\
\Xhline{1.2pt}
\end{tabular}
}
\caption{Architectural comparisons of autoregressive Transformer models and their recall localization.}
\label{Table5}
\end{table*}

\section{Analyzing Factors in Causal Effect Concentration Differences}
\label{Appendix_D}
To investigate differences in the concentration of causal effects, as observed in Figures~\ref{Figure2} and \ref{Figure5}, we present two hypotheses and analyze each in detail.\\
\indent The first hypothesis is that the attention-centered concentration of causal effects observed in Qwen-based models may come from the presence of multi-token subjects. To test this possibility, we repeat the same evaluation procedure under the constraint that subjects consist of a single token.\\
\indent The second hypothesis considers that architectural design factors---such as MLP hidden size, the number of attention heads, and the type of attention mechanism (MHA vs. GQA)---as well as the tokenizer and vocabulary size may influence where causal effects tend to be concentrated within the network. To evaluate this, we conduct a comparative analysis across models with diverse structural configurations.
\subsection{Impact of Subject Token Length on Causal Effect}
We hypothesized that multi-token subjects which consists may contribute to the concentration of causal effects observed in the early Attention layers of Qwen-based models. This hypothesis implies that single-token subjects will not yield the same results as multi-token subjects. To examine this possibility, we reconstruct the \textsc{CounterFact} dataset~\cite{meng2022locating}, which contains multi-token subjects, to be suitable for single-token subject analysis. We paraphrase the original prompts to ensure all models produce the same predictions and ultimately build a dataset of 100 single-token subject sentences for use in causal tracing experiments.\\
\indent Figure~\ref{appendix_figure_single_token} shows the restoration effects under the single-token subject setting. The experimental results confirm that the architecture-specific patterns remain consistent. In GPT-2 XL and LLaMA-3.2-1B, causal effects remain concentrated in the early MLP layers. In contrast, Qwen-2.5-1.5B and DeepSeek-R1-Distill-Qwen-2.5-1.5B continue to exhibit stronger causal concentration in the early Attention layers than GPT-family models.\\
\indent These results indicate that Qwen’s reliance on early Attention layers is not an artifact of multi-token subjects, but rather an intrinsic property of the model. Furthermore, across all models, we observe a consistent increase in causal effect on the relation token that immediately follows the subject, suggesting that in the single-token subject setting, the relation token plays a compensatory role by more prominently anchoring factual retrieval. In multi-token subjects, factual knowledge may accumulate across tokens and become consolidated at the last subject token, which could serve as the entry point for object prediction and the primary locus of factual recall concentration. By contrast, in the single-token subject setting, the subject provides too few token positions for accumulation, so the model may exploit factual knowledge at the subsequent relation token to initiate object prediction, leading factual recall to be concentrated at that position.
\subsection{Impact of Architectural Factors on Causal Effect}
We also investigate whether specific architectural parameters influence the localization of factual recall. Four aspects are examined: the hidden size of the MLP, the number of attention heads, the type of attention mechanism, the tokenizer type, and vocabulary size. A summary of these comparisons is provided in Table~\ref{Table5}.\\
\indent First, comparison across models indicates that MLP hidden size does not consistently determine where recall is localized. For instance, GPT-2 XL (6,400) and LLaMA-3.2-1B (8,192) show recall concentrated in early MLP layers, while Qwen-2.5-1.5B and DeepSeek-R1-Distill-Qwen-1.5B (8,960) exhibit recall centered in early attention layers. However, this trend does not hold for LLaMA-3.1-8B, which has an even larger hidden size of 14,336 but still demonstrates MLP-centered recall.\\
\indent Second, the number of attention heads also shows no consistent relationship with recall localization. Models with very high head counts, such as GPT-2 XL (1,200) and LLaMA-3.1-8B (1,024), rely on MLP layers, whereas Qwen-2.5-1.5B and DeepSeek-R1-Distill-Qwen-1.5B, with only 336 heads, rely on attention layers. Yet, larger models like Qwen-2.5-14B and DeepSeek-R1-Distill-Qwen-14B, with 1,920 heads, still display attention-centered recall, contradicting a simple head-count hypothesis.\\
\indent Third, the type of attention mechanism, whether multi-head attention (MHA) or grouped-query attention (GQA), cannot by itself explain the observed patterns. GPT-2 XL is the only model in this set using MHA and exhibits MLP-centered recall. In contrast, all other models use GQA, but with diverging outcomes: LLaMA models still localize recall in MLP layers, while Qwen models consistently localize recall in early attention layers.\\
\indent Fourth, tokenizer-level properties or vocabulary size may contribute to the observed patterns, but they do not fully account for them. While all models employ BPE-based tokenization, they differ substantially in vocabulary size and structure. For example, GPT-2 XL uses 50,257 tokens, Qwen-2.5-1.5B and DeepSeek-Qwen models use 151,936 tokens, and LLaMA-3.2-1B uses 128,256 tokens. Interestingly, despite LLaMA-3.2-1B having a vocabulary size comparable to that of Qwen, its causal effect distribution aligns more closely with GPT, suggesting that vocabulary size alone does not explain the observed differences.\\
\indent Together, these findings indicate that recall localization cannot be attributed to any single factor. Neither architectural parameters---such as MLP hidden size, attention head count, or attention mechanism type---nor the tokenizer or vocabulary size consistently explain the observed patterns. Instead, recall localization appears to arise from more complex interactions among these design choices, coupled with training objectives and optimization dynamics, which warrant further investigation.
\end{document}